\documentclass[a4paper]{cas-sc}

\usepackage[authoryear]{natbib}
\usepackage{amsmath,amssymb,amsfonts,amsthm}
\usepackage{booktabs}
\usepackage{multirow}
\usepackage{array}
\usepackage{xcolor}
\usepackage{url}
\usepackage{algorithm}
\usepackage{algpseudocode}
\usepackage{adjustbox}

\definecolor{revcolor}{rgb}{0.00,0.25,0.75}
\long\def\rev#1{#1}

\let\arxivPrintCorNotes\printcornotes
\let\arxivPrintAuthorNotes\printfnotes
\renewcommand{\printcornotes}{\arxivPrintCorNotes\arxivPrintAuthorNotes}
\renewcommand{\printfnotes}{}

\ExplSyntaxOn
\cs_set:Npn \__first_footerline:
  {
    \group_begin:
    \small \sffamily
    \ifnum\theblind>0\relax
    \else
      \__short_authors:
    \fi
    \group_end:
  }
\ExplSyntaxOff

\makeatletter
\AtBeginDocument{%
  \@fleqnfalse
  \@mathmargin\@centering
}
\makeatother

\newcommand{\method}{R\nobreakdash-LLMC}

\newcommand{\qwensmall}{Qwen3-4B}
\newcommand{\qwenbig}{Qwen3-14B}
\newcommand{\Nbon}{8}
\newcommand{\Temp}{0.7}
\newcommand{\tauU}{0.75}

\begin{document}
\let\WriteBookmarks\relax
\def\floatpagepagefraction{1}
\def\textpagefraction{.001}
\shorttitle{Value-Guided Language-Model Action Selection}
\shortauthors{Preprint}

\title[mode=title]{Evaluable Language-Model Treatment Policies for Sepsis Resuscitation via Offline Value-Guided Selection}

\begin{abstract}
Offline treatment learning must balance imitation, which stays close to logged
evidence but ignores outcomes, against value maximisation, which uses outcomes
but drifts toward treatments the records cannot verify. Numerical
reinforcement learning strikes this balance during training. For large language models (LLMs), directly using sparse
retrospective outcomes for fine-tuning raises challenges of delayed credit
assignment and limited evidence for alternative treatments. We present R-LLMC, which
strikes the balance at inference for sepsis resuscitation. A
behaviour-supervised LLM, decoded into an exact distribution over 25
vasopressor-fluid actions, proposes clinician-anchored candidates; an offline
critic selects among them by estimated long-term value. The policy has a
closed form for off-policy evaluation, and its rationale is written after
selection, so explanation cannot steer treatment. In experiments on held-out
MIMIC-III patients, R-LLMC exceeded the clinician reference in doubly robust
value by $+0.214$ (95\% CI $0.091$--$0.300$), with fitted Q evaluation
agreeing.
\end{abstract}


\author[1]{Yue Pei}
\fnmark[1]
\ead{peiyue@buaa.edu.cn}

\author[2]{Shining Bo}
\fnmark[1]
\ead{boshining@bjmu.edu.cn}


\author[3]{Yingying Zhang}
\ead{zhangyingying@buaa.edu.cn}


\author[2]{Qinggang Ge}
\ead{qingganggelin@126.com}

\author[2]{Zhiling Zhao}
\cormark[1]
\ead{13439726488@163.com}

\author[4]{Jian Cheng}
\cormark[1]
\ead{jian_cheng@buaa.edu.cn}

\author[3,4]{Haogang Zhu}
\cormark[1]
\ead{haogangzhu@buaa.edu.cn}

\affiliation[1]{organization={School of Artificial Intelligence, Beihang University},
                city={Beijing}, postcode={100083}, country={China}}

\affiliation[2]{organization={Department of Intensive Care Medicine, Peking University Third Hospital},
                city={Beijing}, postcode={100191}, country={China}}


\affiliation[3]{organization={Hangzhou International Innovation Institute, Beihang University},
                city={Hangzhou}, postcode={311115}, country={China}}

\affiliation[4]{organization={School of Computer Science and Engineering, Beihang University},
                city={Beijing}, postcode={100191}, country={China}}

\fntext[fn1]{Equal contribution}
\cortext[cor1]{Corresponding author}

\maketitle

\section{Introduction}
 
Sepsis resuscitation is a sequential treatment
problem~\citep{hotchkiss2016sepsis,singer2016third,evans2021surviving}.
Fluids may restore perfusion while increasing fluid burden, vasopressors
support arterial pressure with competing risks, and each intervention changes
the context of later decisions whose consequences become apparent only as the
patient's course unfolds. Whether a treatment strategy benefits patients is
therefore a property of the policy over the whole course, and retrospective
records reveal outcomes only for treatments that were actually given.
Learning treatment policies from such records faces a well-known
tension~\citep{fujimoto2019off,kumar2020conservative}. Imitating clinicians
keeps recommendations close to logged evidence but ignores outcomes, so it
reproduces practice rather than improving on it. Maximising estimated outcome
value uses that evidence, but estimates for treatments rarely given in
comparable states are extrapolations, and maximisation is drawn toward
precisely the treatments the records cannot verify. A useful aid must
balance the two. It must also be assessable from retrospective records
before prospective use, and it should expose the clinical context behind
each recommendation for review.
 
Existing data-driven approaches address different parts of this problem.
Risk models built on longitudinal EHR trajectories support recognition and
stratification~\citep{shashikumar2021artificial,zhu2025explainable,cissoko2025predicting,li2026hydast},
and workflow-integrated systems bring such estimates closer to care
delivery~\citep{goh2021artificial,adams2022prospective}, but neither
specifies the next treatment. Sequential decision models trained on ICU
records do specify one, using reinforcement learning (RL) to connect a
present action to a delayed
outcome~\citep{komorowski2018artificial,liu2024reinforcement}.
Value-maximising policies such as the AI Clinician sit at the second end of
the tension, ranking candidate actions across the full action space. Offline
RL balances the two ends inside the training objective: conservative
penalties lower the values of rarely logged
actions~\citep{kumar2020conservative}, behaviour-constrained methods restrict
the learned policy toward logged actions~\citep{fujimoto2019off}, and
in-sample methods avoid querying unlogged actions in the value
backup~\citep{kostrikov2021offline}. Because these policies define
probabilities over a fixed treatment space, their value can also be
estimated retrospectively by off-policy evaluation
(OPE)~\citep{thomas2016data,jiang2016doubly,gottesman2019guidelines}. What
they lack is a clinical interface. The patient state is a standardised
numerical vector and the recommendation is a dose index, so the clinical
context supporting a recommendation is not itself exposed for
review~\citep{sokol2025artificial}.
 
Large language models (LLMs) offer that interface. They answer clinical
questions~\citep{singhal2023large,singhal2025toward}, read serialised records
with named clinical fields~\citep{zhou2025large,wu2024large}, and are being
connected to structured treatment
recommendations~\citep{de2025heliot,li2025large}, so their inputs and outputs
are expressed in clinical terms. Direct outcome-driven fine-tuning of LLMs is challenging in this
retrospective setting. The records contain outcomes only for logged treatment
trajectories, leaving the outcomes of alternative recommendations unobserved.
Direct policy-gradient fine-tuning against a terminal outcome faces delayed
credit assignment across clinical decisions and action tokens. A
trajectory-importance-weighted implementation can additionally suffer from
high variance under limited policy--behaviour overlap. These challenges
motivate our use of offline value information for inference-time selection
rather than direct outcome-driven fine-tuning of the proposal model. For LLM
recommenders trained by likelihood or imitation, proposal likelihood alone
does not measure long-horizon treatment
value~\citep{thirunavukarasu2023large,gu2024probabilistic}. Their fluency
compounds the problem. A justification generated together with a
recommendation can be mistaken for evidence of
benefit~\citep{farquhar2024detecting,asgari2025framework,turpin2023language},
and a free-text recommendation is a sample of text rather than an explicit
distribution that OPE could evaluate. Work that combines the two families has
typically kept the policy numerical; MORE-CLEAR, for example, uses a frozen
LLM to encode clinical notes as state features for offline treatment
learning~\citep{lim2026large}. What is missing is a way to strike the
balance for a policy defined by the LLM itself, without outcome gradients,
while keeping that policy explicit enough to evaluate.
 
\begin{figure}
  \centering
  \includegraphics[width=0.99\linewidth]{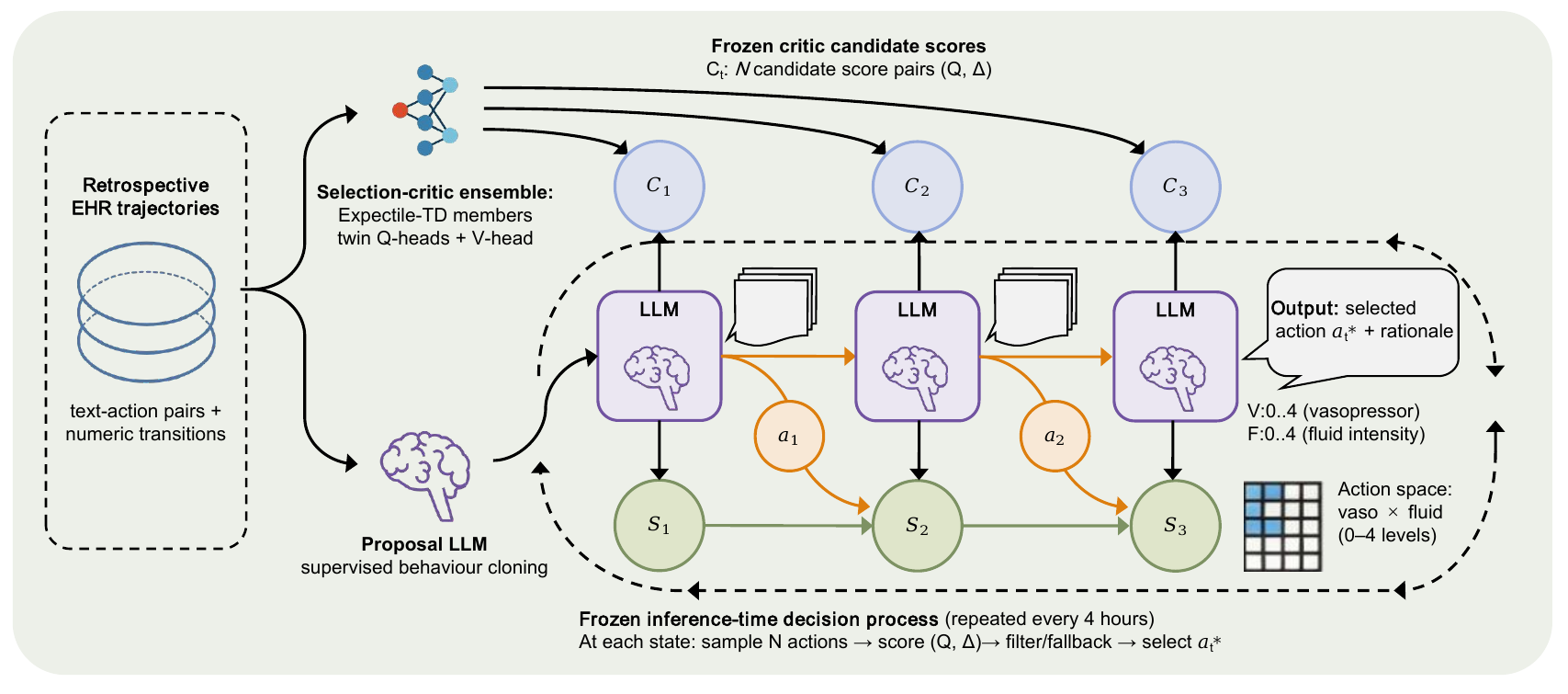}
  \caption{\textbf{R-LLMC training and inference.}
  \rev{Left, retrospective EHR trajectories train the proposal LLM by
  behaviour cloning and the expectile-TD selection critic from terminal
  outcomes. Right, both are frozen at inference: at each 4-hour state $S_t$,
  $N$ actions are sampled from the proposal and ranked by disagreement preference
  and value before the highest-ranked sampled action $a_t^\star$ is
  returned with a post-selection rationale.}}
  \label{fig:method}
\end{figure}

We present \method, which moves the balance from training to inference and
realises it through structure rather than through a regularised objective
(Fig.~\ref{fig:method}). The two ends of the tension are assigned to separate
components, connected by a shared 25-action vasopressor--fluid vocabulary. A
behaviour-supervised LLM reads named clinical fields and emits a structured
action, and constrained factorised decoding over its valid tokens yields the
exact proposal probability of every action; imitation thus determines which
candidates are considered. A frozen numerical critic, trained by expectile
temporal-difference learning on the aligned numerical view of the same
state, then ranks $N$ candidates sampled from this proposal, with a
preference for low twin-head disagreement; value thus decides among them.
Outcome information enters through selection rather than through gradients
to the LLM. Because candidates are drawn from the proposal, selection
reorders clinician-like actions rather than searching the full action space,
and the Best-of-$N$ construction~\citep{beirami2024theoretical} bounds how far
the policy can move from the proposal, making the number of candidates a
single control of the trade-off. This anchoring does not by itself guarantee
logged-data coverage for every selected action. We express the induced
policy, including the disagreement preference, fallback and tie-breaking, in
closed form, which supplies exact target probabilities for OPE. Finally, a
separately trained LLM generates a rationale only after the action is fixed,
so that, unlike a jointly generated justification, the text exposes the
displayed evidence for review without altering the action.
 
Our contributions are threefold.
(1) We formulate an LLM treatment recommender as an explicit, exactly
computable distribution over a clinical action vocabulary through
constrained factorised decoding. This places language policies within the
same offline value-learning and off-policy evaluation framework as numerical
treatment policies.
(2) We realise the imitation--value trade-off for such a policy at inference
rather than in training. A frozen value-guided selection operator admits
delayed-outcome information without outcome-derived gradients to the
proposal model, redistributes probability only within the proposal support,
and bounds the policy's divergence from the proposal. Its induced policy,
including the disagreement preference, fallback and tie-breaking, has a
closed form that supplies exact target probabilities for evaluation.
(3) We evaluate the framework on held-out MIMIC-III patients, where
value-guided selection improved estimated outcome value over both the
supervised proposal and an estimated clinician reference under doubly robust
and fitted Q evaluation. Numerical-proposal and direct-critic controls
attribute this gain to the selection operator, and show that
proposal-constrained selection requires fewer clipped importance ratios than
direct critic maximisation under the reported OPE protocol. A frozen language
proposal was also reused at an independent site with locally fitted
numerical components.

\section{Related work}

\textbf{Sequential treatment learning from retrospective EHRs.}
The AI Clinician learns treatment values over clustered physiological states
and a discrete vasopressor--fluid
grid~\citep{komorowski2018artificial}. Subsequent work studies deep value
functions, continuous state representations, and other ICU
interventions~\citep{liu2024reinforcement,sokol2025artificial}. Retrospective
policy evaluation depends on the cohort, state and action definitions, and
assumptions linking logged trajectories to target-policy
value~\citep{thomas2016data,jiang2016doubly,gottesman2019guidelines}.
\method\ retains a long-horizon treatment objective while constructing the
action policy through language proposals and numerical candidate selection.

\textbf{Clinical language models and treatment interfaces.}
Medical language models support question answering and note-centred
tasks~\citep{singhal2023large,singhal2025toward,thirunavukarasu2023large}, and
can process serialised or longitudinal clinical
records~\citep{zhou2025large,wu2024large}. Their use in decision making requires
attention to temporal grounding, calibration, and reliable structured
output~\citep{gu2024probabilistic,farquhar2024detecting,asgari2025framework}.
MORE-CLEAR uses a frozen LLM to represent clinical notes and combines that
representation with structured variables for offline treatment
learning~\citep{lim2026large}. Here the language model occupies a different
position: it defines the treatment proposal distribution itself, which is
then refined by a numerical selector.

\textbf{Offline value learning and sequence-policy refinement.}
Conservative objectives limit optimistic value extrapolation~\citep{kumar2020conservative},
behaviour-constrained methods restrict policy construction~\citep{fujimoto2019off},
and implicit Q-learning uses an upper-expectile state value fitted at logged
actions~\citep{kostrikov2021offline}. Value-predictor disagreement is also used
as a heuristic for extrapolation sensitivity~\citep{an2021uncertainty}.
Numerical value learning can guide sequence policies at different stages:
QDT uses dynamic programming to relabel return-to-go targets before sequence
policy training~\citep{yamagata2023q}; QT combines Q-value regularisation
during training with value-based selection among return-conditioned candidates
at inference~\citep{hu2024q}. Best-of-$N$ provides a sampling-and-selection
mechanism whose induced policy has been studied
theoretically~\citep{gao2023scaling,beirami2024theoretical}.
\method\ combines an expectile-trained numerical critic with a supervised
language proposal through a shared treatment vocabulary. Its explicit total
action order accounts for disagreement, value, and fallback in the
finite-action Best-of-$N$ distribution.
Unlike QDT and QT, \method\ leaves the training objective of the language
proposal unchanged and introduces value information only through a frozen
selection operator whose induced policy is available in closed form.

\section{Methodology}
\label{sec:method}

\rev{R-LLMC separates action proposal from outcome-based ranking, and gives
each role one of two aligned representations of the same longitudinal EHR
(Fig.~\ref{fig:method}). The language policy reads a clinical textual view and
learns a behaviour-anchored proposal distribution from logged clinician
actions. The numerical value model reads the corresponding standardised
trajectory and estimates long-horizon outcome utility from retrospective
transitions. The two communicate only through a shared code for the discrete
vasopressor--fluid actions.}

\rev{At each decision point the proposal supplies sampled candidates, the
critic scores each of them on the numerical view with a conservative value and
a twin-head disagreement signal, and a deterministic rule prioritises
threshold-passing candidates and returns the highest-ranked one. The
components do not vote: the proposal fixes what is considered, the critic
fixes the order. The explanation is generated only once the action is fixed
and cannot affect selection. Each component serves one of the requirements
set out in Section~1. The proposal turns the LLM's recommendation into an
explicit distribution learned from logged clinician actions. Behaviour
cloning shapes proposal probabilities without imposing an empirical support
mask. The critic contributes delayed-outcome information within the proposal
support. The
selection rule yields a closed-form policy that can be evaluated
retrospectively. The rationale exposes the displayed evidence for review
without entering selection.}

\subsection{Problem formulation}
\label{sec:problem}
We formulate longitudinal sepsis resuscitation as a finite-horizon Markov
decision process
$\mathcal{M}=(\mathcal{S},\mathcal{A},P,R,\gamma)$ on a 4-hour clinical
timescale, \rev{using recorded clinical summaries as an approximation of the
patient state}. A trajectory is
$\zeta=(s_0,a_0,r_0,\ldots,s_{T-1},a_{T-1},r_{T-1},s_T)$, where
$s_t\in\mathcal{S}$ is \rev{the recorded clinical summary at decision time
$t$}. The action $a_t\in\mathcal{A}$ is a haemodynamic treatment. The unknown
patient dynamics $P$ are represented only through retrospective transitions.
The objective of a treatment policy $\pi$ is
\begin{equation}
J(\pi)
=\mathbb{E}_{\zeta\sim(P,\pi)}
\left[\sum_{t=0}^{T-1}\gamma^t r_t\right].
\label{eq:objective}
\end{equation}
Thus, the policy objective concerns expected long-horizon patient outcome rather
than the local plausibility of an individual action proposal.

\rev{All learning takes place on logged data. We write the available
transitions as
$\mathcal{D}=\{(s_t,a_t,r_t,s_{t+1},d_t)\}$, where $d_t$ marks a transition
that reaches a terminal state. Two properties of $\mathcal{D}$ shape every
design choice below: the reward is observed only at the end of a stay, and
it is observed only for the treatment that was actually given. The outcomes
of alternative actions at that decision point are therefore not observed.}

The agent operates at two coupled timescales. At the \emph{decision level},
one action is chosen every 4 hours and its consequences can unfold over
the rest of the ICU trajectory. At the \emph{generation level}, an action
is expressed by an autoregressive token sequence. Optimising the language
model directly with the terminal outcome would nest the second horizon inside
the first. If $w_{t,1:L_t}$ denotes the tokens used to express decision $t$, a
sequence-level policy-gradient estimator would contain a term of the form
\begin{equation}
\nabla_\theta J
=\mathbb{E}\!\left[
(G-b)\sum_{t=0}^{T-1}\sum_{\ell=1}^{L_t}
\nabla_\theta\log\pi_\theta
\big(w_{t,\ell}\mid w_{t,<\ell},o_t\big)\right],
\label{eq:tokenpg}
\end{equation}
in which a single delayed and sparse return $G$ must be credited across every
clinical decision of the stay and across every token used to express each of
them. \rev{A compact action string shortens the inner sum but leaves the
decision-level credit assignment unchanged. A trajectory-importance-weighted estimate of the expectation in
Eq.~\ref{eq:tokenpg} would involve products of importance ratios over the
stay, which can become concentrated on a few trajectories
(Appendix~\ref{app:ope}). Such an estimator can have high variance under
limited policy--behaviour overlap, motivating our choice to introduce outcome
information through a separate numerical critic rather than directly
fine-tuning the language proposal. If one
model produced both the action and its explanation, such fine-tuning would
also move the two distributions together.}

\method\ instead performs \rev{a single step of critic-guided policy
improvement, in the spirit of generalised policy iteration and of approaches
that connect sequence policies to numerical value
learning~\citep{hu2024q,yamagata2023q}}. A supervised language proposal
policy $\pi_\theta^{\mathrm{prop}}$ is fitted by behaviour cloning on
clinician actions and defines the behaviour-anchored candidate distribution.
An independent numerical selection critic $\mathcal{E}_\omega$ estimates the
downstream utility of sampled candidates from logged trajectories. A
non-parametric operator $\mathcal{I}_{N,\xi}$ then uses these distinct outputs
to improve the proposal at inference:
\begin{equation}
\pi^{\text{R-LLMC}}_{N,\xi}
=\mathcal{I}_{N,\xi}
\big[\pi_\theta^{\mathrm{prop}},\mathcal{E}_\omega\big].
\label{eq:gpi}
\end{equation}
\rev{No outcome-derived gradient reaches $\theta$, so the sum over tokens in
Eq.~\ref{eq:tokenpg} never has to be formed. Outcome information enters only
through the critic and the selection operator, at the timescale on which the
outcome is defined.}

\subsection{Dual-view clinical action interface}
\label{sec:stateaction}
\rev{The proposal and critic require different representations, but their
outputs must refer to the same treatment decision. This section defines the
shared interface that makes this possible.}

\textbf{A dual-view clinical state.}
The same recorded summary $s_t$ is exposed to the two components through
different but deterministically aligned representations,
\begin{equation}
x_t=g_{\mathrm{num}}(s_t)\in\mathbb{R}^{d},
\qquad
o_t=g_{\mathrm{text}}(s_t)\in\mathcal{O}.
\label{eq:stateviews}
\end{equation}
The numerical state $x_t$ contains the standardised values of 20 variables,
including vital signs, laboratory measurements, SOFA, GCS, and treatment
context accumulated through the current row (maximum vasopressor exposure and
cumulative intravenous fluid). It is the sole input to the selection critic.
The textual observation $o_t$ serialises the corresponding unstandardised
measurements into semantically labelled fields, for example
\texttt{<HeartRate:74, normal> | <MeanBP:76, normal> |
<Lactate:1.1, normal> | \dots}. \rev{Proposals are formed from the current
observation only ($H_{\mathrm{prop}}=1$). Accumulated treatment context is
already part of that observation, which keeps the language input compact.
The rationale may additionally draw on up to $H_{\mathrm{rat}}\leq4$
immediately preceding observations of the same trajectory, so that a stated
trend can be checked against the values the clinician is shown.} Both views
originate from the same patient and time row and share an identical action
code; they are aligned representations of the same longitudinal EHR, not
independent raw clinical modalities. \rev{Because the two views carry the
same measurements, a difference between proposals built on them reflects the
model class rather than the information available; the numerical-proposal
control of Appendix~\ref{app:controls} relies on this. The textual view keeps
the language policy's input and output in clinical terms, which is what makes
its recommendations reviewable. The standardised numerical view gives the
critic a stable, low-dimensional input for temporal-difference learning and
prevents it from scoring unconstrained free text.}

\textbf{Treatment action and language outputs.}
The environment action is
\begin{equation}
a_t=(v_t,f_t)\in\mathcal{A}
=\{0,\ldots,4\}\times\{0,\ldots,4\},
\qquad
\iota(a_t)=5f_t+v_t\in\{0,\ldots,24\},
\label{eq:action}
\end{equation}
where $v_t$ and $f_t$ are the vasopressor and intravenous-fluid intensity
bins for the next 4-hour interval. For each treatment dimension, bin 0
denotes zero exposure. Non-zero doses are partitioned into bins 1--4
at the 25th, 50th, and 75th percentiles of the non-zero empirical
dose distribution among training decision-target rows. The compact
$5\times5$ grid balances dosing resolution with the number of logged
observations per action bin. \rev{Discretisation does not ensure adequate
state-specific coverage for every action.} The fixed index $\iota$
is the shared action interface between textual proposals and numerical critic
outputs. \rev{Internal selection and evaluation use this 25-action joint
space. At the independent site, the proposal is marginalised over fluid
before selection and evaluation over five vasopressor levels.}

\textbf{Structured proposal and post-selection rationale.}
The language proposal policy must return a valid, inspectable action
distribution rather than an unconstrained free-text recommendation. It is
supervised to emit the compact target
\texttt{V:<0-4> F:<0-4>}. At inference, the policy does not rely on
unconstrained free-text sampling or on rejecting malformed completions.
The next-token distributions are instead restricted to the five valid dose
bins and all five fluid conditionals are evaluated, yielding the exact
normalised proposal
\begin{equation}
p_{\theta,T_{\mathrm{s}}}(v,f\mid o_t)
=p_{\theta,T_{\mathrm{s}}}(v\mid o_t)\,
p_{\theta,T_{\mathrm{s}}}(f\mid o_t,v),
\qquad (v,f)\in\mathcal{A}.
\label{eq:parsedpolicy}
\end{equation}
Each factor is normalised over $\{0,\ldots,4\}$, so
Eq.~\ref{eq:parsedpolicy} defines a proper distribution over all 25 actions;
\rev{$T_{\mathrm{s}}$ is the sampling temperature applied to the restricted
logits.} \rev{The proposal probability of every action, including actions the
language policy would rarely emit, is therefore available in closed form, and
the policy does not depend on whether a generated string can be parsed. The
factorisation is exact by the chain rule and requires only a fixed, small
number of next-token evaluations per decision. Because the same decoding
applies to any backbone, including one without fine-tuning, every language
policy in Section~4 is evaluated as an explicit distribution. The temperature
controls how concentrated the proposal is and hence how many distinct
candidates Best-of-$N$ sampling considers.}

Selection and explanation use separate language outputs. After the selector
fixes $a_t^\star$, the rationale model produces an
evidence-linked justification from the past-only clinical history
$h_t=(o_{\max(0,t-H_{\mathrm{rat}}+1)},\ldots,o_t)$,
$1\leq H_{\mathrm{rat}}\leq4$:
\begin{equation}
\rho_t^\star\sim
\pi_\varphi^{\mathrm{rat}}
\bigl(\cdot\mid h_t,a_t^\star\bigr).
\label{eq:rationalise}
\end{equation}
\rev{Conditioning on the already selected action, rather than generating the
action and its justification together, keeps the explanation outside the
selection pathway. No property of the generated text can alter the fixed
recommendation, and the rationale is not selection evidence.} The complete
clinician-facing response $z_t=(a_t,\rho_t)$ therefore factorises as
\begin{equation}
p(z_t\mid x_t,o_t,h_t)
=\pi_{N,\xi}^{\mathrm{R\mbox{-}LLMC}}
(a_t\mid x_t,o_t)\,
\pi_\varphi^{\mathrm{rat}}(\rho_t\mid h_t,a_t),
\label{eq:response}
\end{equation}
where only $a_t$ enters the clinical transition and the outcome objective.
The rationale is a reviewable, post-selection justification. We do not treat
it as an account of the computation that produced the action, as evidence used
to select the action, or as a guarantee of clinical correctness.

\textbf{Supervised language interface.}
The proposal and rationale outputs require separate supervision because only
the proposal distribution enters action selection and evaluation. Let
$\mathcal{D}_{\mathrm{prop}}
=\{(u^{\mathrm{prop}}(o_t),y_t^{\mathrm{act}})\}$
contain textual observations paired with logged structured actions, and
let
$\mathcal{D}_{\mathrm{rat}}
=\{(u^{\mathrm{rat}}(h_t,a_t),\rho_t)\}$ contain past-only histories and the
corresponding logged clinician actions paired with structured rationale
targets.
\rev{The proposal and rationale modules are fitted on the two supervision
sets with completion-only objectives,}
\begin{equation}
\mathcal{L}_{m}(\vartheta_m)
=-\,\mathbb{E}_{(u,y)\sim\mathcal{D}_{m}}
\sum_{\ell=1}^{|y|}
\log\pi_{\vartheta_m}\!\big(y_\ell\mid y_{<\ell},u\big),
\qquad
m\in\{\mathrm{prop},\mathrm{rat}\}.
\label{eq:sft}
\end{equation}
\rev{The proposal parameters $\vartheta_{\mathrm{prop}}=\theta$ and the
rationale parameters $\vartheta_{\mathrm{rat}}=\varphi$ are fitted separately,
so that supervision on explanatory style cannot perturb the proposal
distribution used by the selector. Neither objective includes an
outcome-based loss.}
Rationale supervision was conditioned on past-only histories and the
corresponding logged clinician actions; no quantity computed by the selection
critic entered rationale training, whereas at inference the model receives the
action already fixed by the selector. \rev{Because the selector chooses only
among actions sampled from a proposal fitted to clinician behaviour, the
actions that condition the rationale at inference resemble the logged actions
it was trained on.}

\textbf{Terminal outcome reward.}
The value model needs a target defined on the patient's later trajectory
rather than on the likelihood of a proposal. Let the last transition of a
trajectory arrive at the recorded hospital outcome.
The raw and optimisation rewards are
\begin{equation}
\begin{aligned}
r^{\mathrm{raw}}_{t}&=
\begin{cases}
0, & s_{t+1}\ \text{is non-terminal},\\
+100, & s_{t+1}\ \text{is terminal and the patient survives},\\
-100, & s_{t+1}\ \text{is terminal and the patient dies},
\end{cases}\\
r_t&=0.01\,r^{\mathrm{raw}}_t,
&G_t&=\sum_{k=t}^{T-1}\gamma^{\,k-t}r_k .
\end{aligned}
\label{eq:reward}
\end{equation}
The corresponding action value,
\begin{equation}
Q^\pi(x_t,a_t)
=\mathbb{E}_{\pi,P}\!\left[G_t\mid x_t,a_t\right],
\qquad
V^\pi(x_t)
=\mathbb{E}_{a\sim\pi(\cdot\mid x_t)}[Q^\pi(x_t,a)],
\label{eq:valueobjective}
\end{equation}
is the numerical signal the agent uses to rank candidate treatments. Using
only the recorded hospital outcome avoids fixing trade-offs between
physiological targets in advance, at the cost of a signal that is sparse and
arrives late; Section~\ref{sec:value} propagates it backwards along the
numerical trajectory. \rev{With $\gamma=0.99$ and terminal label
$Y\in\{-1,+1\}$, the initial return is $G_0=\gamma^{T-1}Y\in[-1,1]$, so all
reported values are discounted outcome utilities on this scale.}

\subsection{Outcome-aligned ensemble critic}
\label{sec:value}
\rev{The selector needs a numerical ranking signal for each proposed
candidate, but the record contains an outcome only for the action that was
given. Estimating the others offline invites two failure modes: a backup that
maximises over all treatments assigns unchecked optimistic values to
treatments seldom given in comparable states, and a single fitted network
carries seed-specific error that is invisible at selection time. The selection
critic $\mathcal{E}_\omega(x,a)$ of Eq.~\ref{eq:gpi} addresses both by keeping
the backup within logged actions and returning a twin-head disagreement
heuristic alongside each value.}

We train the selection critic on the numerical view of the logged transitions,
$\{(x_t,a_t,r_t,x_{t+1},d_t)\}$, using $M=3$ independently initialised
members. Each member has a numerical-state encoder, twin action-value heads
$Q_{m,1}$ and $Q_{m,2}$ over the 25 actions, and a scalar state-value head
$V_m$.

\textbf{In-sample state-value evaluation.}
Offline maximisation over all actions can assign spuriously high values
to actions poorly represented in the data. We instead regress the state
value to an upper expectile of target-network action values at
\emph{logged} actions. For critic member $m$, let superscript ${-}$ denote
its slowly updated target network and define
$L_\tau(z)=|\tau-\mathbf{1}[z<0]|z^2$ for
$\tau\in(\tfrac12,1)$. The state-value loss is
\begin{equation}
\delta_{t,m}
=\min_{j\in\{1,2\}}Q^{-}_{m,j}(x_t,a_t)-V_m(x_t),
\qquad
\mathcal{L}^{(m)}_V
=\mathbb{E}_{\mathcal{D}}\!\left[L_\tau(\delta_{t,m})\right].
\label{eq:valueloss}
\end{equation}
\rev{The upper expectile summarises the upper range of target values associated
with logged actions. This avoids maximisation over unobserved actions in the
state-value backup, while retaining a continuation signal for subsequent
action-value learning. As $\tau\to1$, $V_m$ approaches the maximum of the
action value over actions with support in the data at that
state~\citep{kostrikov2021offline}; an intermediate $\tau$ trades this
optimism against sensitivity to noisy targets. Taking the smaller head inside
$\delta_{t,m}$ further limits overestimation of the target.}

\textbf{Outcome propagation to action values.}
The sparse terminal outcome must be propagated to earlier treatment decisions
before candidates can be ranked. The action-value heads use a one-step target
bootstrapped through the target state value,
\begin{equation}
\hat y_{t,m}
=r_t+(1-d_t)\gamma
V^{-}_m(x_{t+1}),
\qquad
\mathcal{L}^{(m)}_Q
=\mathbb{E}_{\mathcal{D}}\!\left[
\frac{1}{2}\sum_{j=1}^{2}
\ell_\beta\!\left(
Q_{m,j}(x_t,a_t)-\hat y_{t,m}\right)\right],
\label{eq:qloss}
\end{equation}
where $\ell_\beta$ is the Huber loss. \rev{Applied repeatedly along a
trajectory, this backup propagates the single outcome recorded at discharge
into an estimated long-horizon score for each treatment decision. Training
remains restricted to actions in the empirical distribution.} Each member
independently minimises $\mathcal{L}^{(m)}_V+\mathcal{L}^{(m)}_Q$.
\rev{Although each head outputs values for all 25 actions, only logged
actions receive regression targets, so values for treatments rarely given in
comparable states are extrapolations of the shared encoder. For this reason
the critic ranks candidates drawn from a clinician-anchored proposal rather
than maximising over the full action space (Section~\ref{sec:select};
Appendix~\ref{app:controls}).}

\textbf{Disagreement-aware evaluation.}
Candidate ranking also requires a conservative summary of the fitted ensemble
and an explicit signal for head disagreement. The two heads are first averaged
across ensemble members,
\begin{equation}
Q^{\mathrm{ens}}_j(x,a)
=\frac{1}{M}\sum_{m=1}^{M}Q_{m,j}(x,a),
\qquad j\in\{1,2\}.
\label{eq:ensembleq}
\end{equation}
Only then does the frozen selection critic form the conservative value and
disagreement returned to the selector:
\begin{equation}
\begin{aligned}
\mathcal{E}_\omega(x,a)
&=\bigl(\bar Q_\omega(x,a),\Delta_\omega(x,a)\bigr),\\
\bar Q_\omega(x,a)
&=\min_{j\in\{1,2\}}Q^{\mathrm{ens}}_j(x,a),\\
\Delta_\omega(x,a)
&=\left|Q^{\mathrm{ens}}_1(x,a)-Q^{\mathrm{ens}}_2(x,a)\right|.
\end{aligned}
\label{eq:evaluator}
\end{equation}
Here $\omega$ denotes the complete ensemble. \rev{Averaging across members
before taking the minimum first reduces seed-specific error and then applies
pessimism, so that $\bar Q_\omega$ and $\Delta_\omega$ reflect divergence
between heads rather than between individual members. Taking the smaller of
the two heads is intended to limit overestimation. Because the twin heads
share targets but differ in initialisation, their gap tends to widen where
the data constrain them weakly, a property used as a heuristic for
extrapolation sensitivity~\citep{an2021uncertainty}. We therefore use
$\Delta_\omega$ only to order candidates, and we fix its cutoff without any
access to the test cohort:}
\begin{equation}
\xi
=\operatorname{Quantile}_{\tau_u}
\left\{\Delta_\omega(x,a):
(x,a)\ \text{is a logged validation pair}\right\}.
\label{eq:xi}
\end{equation}
\rev{The gap is a ranking signal, not ensemble variance, calibrated epistemic
uncertainty, or a clinical-safety probability.} Because the expectile backup
estimates a high-value continuation within the logged action support, this
critic is also not an estimate of $Q^{\pi_\theta^{\mathrm{prop}}}$; its role is
to score candidates for inference-time improvement, and it is kept separate
from every outcome model used later for evaluation.

\subsection{Disagreement-aware Best-of-\(N\) selection}
\label{sec:select}
\rev{What remains is an explicit rule that turns proposal sampling and
numerical ranking into a single recommendation, while keeping the induced
target policy in a form that can be written down and evaluated.}
Let $p_t(a)=p_{\theta,T_{\mathrm{s}}}(a\mid o_t)$ from
Eq.~\ref{eq:parsedpolicy} and draw
$A_t^{(1)},\ldots,A_t^{(N)}\stackrel{\mathrm{iid}}{\sim}p_t$.
Define the low-disagreement action set
\begin{equation}
\mathcal{F}_t
=\{a\in\mathcal{A}:\Delta_\omega(x_t,a)\leq\xi\}.
\label{eq:feasible}
\end{equation}
If at least one sampled action lies in $\mathcal{F}_t$, only sampled
threshold-passing actions remain; otherwise the selector falls back to all
sampled actions. The resulting candidate pool is
\begin{equation}
\mathcal{C}_t
=
\begin{cases}
\{A_t^{(k)}:A_t^{(k)}\in\mathcal{F}_t\},
& \text{if this set is non-empty},\\
\{A_t^{(1)},\ldots,A_t^{(N)}\},
& \text{otherwise},
\end{cases}
\label{eq:filter}
\end{equation}
and the highest conservative value in this pool is returned as the agent's
treatment recommendation:
\begin{equation}
a_t^\star
=\underset{a\in\mathcal{C}_t}{\arg\max}\;
\bar Q_\omega(x_t,a).
\label{eq:select}
\end{equation}
Ties are broken by the \rev{smaller} fixed action index $\iota(a)$. \rev{The
order of the two steps carries the design intent. Filtering before maximising
prioritises sampled candidates below the validation-defined disagreement
threshold before their conservative values are compared. This reduces the
influence of a high value paired with greater head disagreement. The fallback
keeps the policy defined at every state, at the cost of making the threshold
a preference over candidates rather than a hard support or safety
constraint.} The value $\bar Q_\omega$ is queried only on $(x_t,a)$ pairs in
the numerical action space, never on generated text. The rationale of
Eq.~\ref{eq:rationalise} is produced only after the recommended action is
fixed and is never used as selection evidence.

\textbf{Exact policy induced by filtered Best-of-\(N\).}
\rev{Sampling makes the recommendation stochastic. If the agent's action
distribution were available only through simulation, evaluation against logged
data would require a sampling approximation. The procedure above avoids this
by inducing a distribution over the 25 actions that can be written in closed
form.} To expose it, define the disagreement--value rank
\begin{equation}
c_t(a)
=\left(
\mathbf{1}\!\left[\Delta_\omega(x_t,a)\leq\xi\right],
\bar Q_\omega(x_t,a),
-\iota(a)
\right),
\label{eq:rank}
\end{equation}
ordered lexicographically. Maximising $c_t$ over the $N$ samples is
exactly equivalent to filtering first, falling back if necessary, and then
maximising value with deterministic tie-breaking. Let the lower cumulative
proposal masses under this total order be
\begin{equation}
F_t(a)=\sum_{b:c_t(b)\leq c_t(a)}p_t(b),
\qquad
F_t^{-}(a)=\sum_{b:c_t(b)<c_t(a)}p_t(b).
\label{eq:boncdf}
\end{equation}
The probability that $a$ is the best-ranked member of $N$ independent
proposals is therefore
\begin{equation}
\boxed{
\pi_{N,\xi}^{\text{R-LLMC}}(a\mid x_t,o_t)
=F_t(a)^N-\big(F_t^{-}(a)\big)^N .
}
\label{eq:inducedpolicy}
\end{equation}
This finite-difference expression gives the complete 25-action target policy
used for evaluation without a Monte Carlo approximation. The distribution
normalises to one, reduces to $p_t$ at $N=1$, and redistributes mass only
within the proposal support at that decision point, rather than prohibiting a
treatment across states. \rev{An action with $p_t(a)=0$ has
$F_t(a)=F_t^{-}(a)$ and therefore receives no probability, and as $N$ grows
the policy concentrates on the highest-ranked action among those the proposal
supports. The operator depends on the proposal only through $p_t$, so any
proposal that defines explicit action probabilities can be substituted
without changing the selector or the evaluation; Appendix~\ref{app:controls}
uses this to replace the language proposal with a numerical one. The same
probabilities serve as OPE target probabilities and allow treatment
allocation to be read directly across clinical states
(Fig.~\ref{fig:heatmap}).}

\rev{\textbf{Departure from the proposal.}
Because Eq.~\ref{eq:rank} defines a total order, Eq.~\ref{eq:inducedpolicy}
is a Best-of-$N$ policy over that order, and its divergence from the
clinician-anchored proposal is bounded by
\begin{equation}
\mathrm{KL}\!\left(\pi_{N,\xi}^{\text{R-LLMC}}(\cdot\mid x_t,o_t)\,\middle\|\,p_t\right)
\leq\log N-\frac{N-1}{N}
\label{eq:klbound}
\end{equation}
\citep{beirami2024theoretical}, about $1.2$ nats at $N=8$, independently of
the critic. $N$ is therefore a single, interpretable control of how far
outcome information may move the policy away from the proposal distribution.
This bound is relative to $p_t$, not to the fitted behaviour reference
$\mu$, and does not guarantee empirical action coverage. By contrast,
maximising $\bar Q_\omega$ over all actions is not constrained by the
proposal at all.}

Because the frozen selection critic is an offline approximation rather than
the exact $Q$-function of the proposal, we describe this operation as
\emph{critic-guided} policy improvement and assess its policy value
separately using DR and FQE. Algorithm~\ref{alg:main} summarises the complete
decision-support agent.

\begin{algorithm}[t]
\caption{R-LLMC: offline training and frozen inference}
\label{alg:main}
\begin{algorithmic}[1]
\Require aligned trajectories $\{(o_t,x_t,a_t,r_t)\}$; rationale histories
$h_t$; proposal count $N$; temperature $T_{\mathrm{s}}$; validation quantile
$\tau_u$
\Statex \textbf{Offline training}
\State Fit proposal $\theta$ by behaviour cloning on logged actions
\State Fit $M{=}3$ expectile-TD critics, average corresponding heads across members, freeze
\State $\xi\gets\operatorname{Quantile}_{\tau_u}\{\Delta_\omega\}$ on logged validation pairs
\State Fit rationale parameters $\varphi$ on logged-action-conditioned targets
\Statex \textbf{Inference at $(o_t,x_t)$, repeated every 4\,h}
\State $p_t\gets p_{\theta,T_{\mathrm{s}}}(\cdot\mid o_t)$; draw $A^{(1:N)}$ independently with replacement from $p_t$
\State $(\bar Q_k,\Delta_k)\gets\mathcal{E}_\omega(x_t,A^{(k)})$ for $k=1,\dots,N$
\State $\mathcal{I}\gets\{k:\Delta_k\le\xi\}$; if empty, $\mathcal{I}\gets\{1,\dots,N\}$
\State $a_t^\star\gets A^{(k^\star)}$, $k^\star=\arg\max_{k\in\mathcal{I}}\bar Q_k$, ties by smaller $\iota$
\State $\rho_t^\star\sim\pi_\varphi^{\mathrm{rat}}(\cdot\mid h_t,a_t^\star)$
\State \Return $a_t^\star$ and $\rho_t^\star$ for clinician review
\end{algorithmic}
\end{algorithm}

\section{Experiments}

The experiments address four questions that follow from the design in
Section~\ref{sec:method}.
(Q1) Does value-guided selection over an explicit LLM proposal yield higher
estimated outcome value than the clinician reference on held-out patients?
(Q2) Which elements produce the estimated gain: behavioural supervision,
value-guided selection, the disagreement preference, the proposal model
class, or the backbone?
(Q3) Can a frozen language proposal be reused at an independent site when
only the numerical components are refitted?
(Q4) What does the explicit action distribution make inspectable, and how
does the post-selection rationale relate to the displayed physiology and the
selected action?

\subsection{Experimental setup}

\textbf{Data sources and cohorts.}
We conducted a retrospective study using preprocessed trajectories derived
from MIMIC-III~\citep{johnson2016mimic}, with the primary analysis restricted
to adult ICU stays meeting the Sepsis-3 definition~\citep{singer2016third}.
The existing patient-level development/test split contained 4,493/500
trajectories. We reserved 450 development trajectories for validation by
stratified sampling on hospital mortality, leaving 4,043 for training.
Preprocessing statistics, model fitting, early stopping, and hyperparameter
selection used only the training and validation partitions; the 500-patient
test set was reserved for final evaluation.

The independent-site analysis used a separate cohort from the ICU of
Peking University Third Hospital. Sepsis status was determined from the
site-provided sepsis indicator, and trajectory-level eligibility and
temporal-coverage criteria yielded 571 patients. Table~\ref{tab:cohort}
summarises the analysed cohorts.

\begin{table}[t]
\caption{\textbf{Analysed trajectory cohorts and splits.} Mortality is
computed from the terminal outcome of each patient trajectory.}
\label{tab:cohort}
\centering
\begin{adjustbox}{max width=\linewidth}
\begin{tabular}{@{}l r r r r@{}}
\toprule
Cohort / split & Patients & Rows & Transitions & Mortality (\%) \\
\midrule
MIMIC-III training   & 4,043 & 43,468 & 39,425 & 15.16 \\
MIMIC-III validation &   450 &  4,882 &  4,432 & 15.11 \\
MIMIC-III test       &   500 &  5,287 &  4,787 & 15.20 \\
Independent-site adaptation & 571 & 9,221 & 8,650 & 34.15 \\
\bottomrule
\end{tabular}
\end{adjustbox}
\end{table}

\textbf{Preprocessing.}
Internal trajectories used a 4-hour grid and the 20 physiological and
treatment-context variables listed in Appendix~\ref{app:data}. Missing
measurements were imputed by past-only within-trajectory forward filling,
followed by training-partition medians for unresolved entries, and
training-set standardisation was frozen for validation and test. Vasopressor
and fluid exposures defined the $5\times5$ action grid of
Eq.~\ref{eq:action}, with zero exposure as level 0 and training-set quartiles
of positive exposures as levels 1--4. At row $j$, state variables stop at $j$
and the action describes treatment in the next interval. The terminal outcome
was scaled to $\{-1,+1\}$ and assigned on arrival at the terminal state.

\textbf{Compared policies.}
In the internal experiments, every compared policy is expressed as a
distribution over the same 25-action grid for selection and OPE.
\textbf{Random} assigns uniform probability to the 25 actions.
\textbf{Clinician behaviour} is an estimated policy $\mu(a\mid x)$ fitted
to logged actions by behaviour cloning and serves as the paired-comparison
reference. \textbf{Neural SARSA-Q} follows the AI-Clinician
paradigm~\citep{komorowski2018artificial} with a continuous-state
$Q$-network and SARSA updates bootstrapping on the next logged action,
rather than clustered states; its Boltzmann temperature is selected on
validation data to match the clinician model's mean action entropy.
\textbf{Base LLM} uses the backbone without fine-tuning and the same
constrained decoding interface ($N=1$).
\textbf{SFT proposal} samples the behaviour-supervised language proposal
directly ($N=1$), isolating behavioural supervision from selection.
\textbf{\method} applies Best-of-8 selection with the disagreement
preference; unfiltered and $N=4$ variants appear in the component analyses.
Appendix~\ref{app:controls} also evaluates the same selector with a numerical
behaviour-cloning proposal and the selection critic used directly as a
greedy or Boltzmann policy.

\textbf{Evaluation.}
\label{sec:ope}
Policy values are estimated from retrospective trajectories by off-policy
evaluation (OPE)~\citep{jiang2016doubly}. The primary estimator is a
self-normalised, stepwise weighted doubly robust (DR)
estimator~\citep{thomas2016data}, which combines outcome regression with an
importance-weighted correction. For each target policy, three independently
seeded fitted Q evaluation (FQE) models were trained on the training
transitions, selected by validation TD loss, and frozen before test
prediction~\citep{le2019batch,hao2021bootstrapping}. Their averaged action
values enter both DR and the complementary FQE-direct estimate. The
evaluation models are fitted separately from the selection critic of
Section~\ref{sec:value}. DR uses a per-step log-ratio cap of $\log20$ in
magnitude. Percentile 95\% intervals come from $B=500$ outcome-stratified
patient-bootstrap resamples conditional on the fitted models, and differences
from the clinician reference use matched resamples. Because DR and FQE-direct
fail in different ways (importance-weight variance versus regression
extrapolation), we treat agreement between them as more informative than
either estimate alone. Appendix~\ref{app:ope} gives the estimator, resampling
definitions, and weight diagnostics.

\textbf{Implementation.}
\label{sec:impl}
The primary experiments use a BF16 \qwensmall\ backbone, with proposal and
rationale parameters fitted separately~\citep{yang2025qwen3}. The
backbone-scale analysis repeats the action-policy pipeline with a separately
fitted \qwenbig\ proposal while sharing all backbone-independent components.
The selection critic is a three-member ensemble of compact multilayer
networks, each with twin action-value heads and a state-value head. Unless
stated otherwise, the proposal uses the current observation and the
rationale model at most four consecutive observations. Outcome propagation
uses $\gamma=0.99$ and expectile $\tau=0.8$. The selector draws $N=\Nbon$
candidates at sampling temperature $T_{\mathrm{s}}=\Temp$, and the
disagreement cutoff $\xi$ of Eq.~\ref{eq:xi} is the $\tauU$ quantile of
disagreement on logged validation pairs. Remaining settings are given in
Appendix~\ref{app:hparams}.

\textbf{Independent-site adaptation protocol.}
To test whether a language proposal can be reused, the MIMIC-III proposal
remained frozen and retained source-domain input processing. Numerical
preprocessing, selection critics, behaviour models, action thresholds,
overlap calibration, and evaluation models were fitted locally. The joint
proposal was first marginalised over fluid,
$p_t^{\mathrm{site}}(v)=\sum_{f=0}^{4}p_t(v,f)$, and selection and OPE then
operated over five vasopressor levels. All site policies were evaluated in
this five-level action space.
Patient-level, outcome-stratified three-fold cross-fitting fitted local
components on two folds and applied them to the held-out fold. A common
overlap constraint, calibrated within each outer-training partition, was
applied to all non-clinician target policies, and both DR importance ratios
and FQE policy expectations used the constrained probabilities. Pooled
out-of-fold predictions were evaluated with $B=500$ patient bootstraps
stratified jointly by outer fold and terminal outcome, holding fitted models
fixed. Appendix~\ref{app:site_overlap} defines the probability transformation
and fold-local parameter-selection rule.

\subsection{Held-out policy value}
\label{sec:results}
We compared all policies on the 500 held-out MIMIC-III patients under DR and
FQE-direct, with paired DR differences from the clinician reference
(Table~\ref{tab:main}, Panel A).
Among the policies in Table~\ref{tab:main}, Panel A, \method\ was the only
policy whose estimated value exceeded the clinician reference under both
estimators, with a paired DR difference of $+0.214$
(95\% interval $[+0.091,+0.300]$). The base LLM fell below the clinician
reference under both estimators. Random and Neural SARSA-Q had negative DR
but positive FQE-direct estimates.
Behavioural supervision encourages clinician-like proposals. Selection
redistributes probability within the proposal support, but does not guarantee
adequate logged-data coverage for the selected actions. The critic ranks
sampled candidates by estimated continuation value, introducing outcome
information at selection. The sign disagreement between DR and FQE-direct
for Random and Neural SARSA-Q highlights sensitivity to importance weighting
and outcome regression; it does not establish the reliability of the
critic rankings. The
base LLM illustrates the motivating problem: an LLM used directly as a
recommender, without behavioural or outcome grounding, is a poorly
performing policy, and the interface makes that deficit measurable on the
same scale as conventional policies. These are retrospective estimates;
trajectory-weight ESS is low for every policy that departs from the
estimated behaviour model, and Appendix~\ref{app:ope} discusses the
implications.

\begin{table*}[t]
\caption{\textbf{Retrospective policy-value estimates with Qwen3-4B in the held-out
MIMIC-III and independent-site cohorts.} Panel A reports full-test-set
estimates in MIMIC-III using 25 joint vasopressor--fluid actions. Panel B
reports the same policy families together with
the SFT proposal using pooled out-of-fold estimates from patient-level
three-fold site adaptation in the independent ICU cohort. Joint proposals
were marginalised over fluid before selection and evaluation over five
vasopressor levels; the proposal LLM was
frozen, and the site-specific value-selection and OPE components were fitted
only within the corresponding outer-training folds. All non-clinician policies
in Panel B used the same fold-local overlap constraint. Intervals are 95\% conditional patient-bootstrap intervals from
$B=500$ outcome-stratified resamples, additionally stratified by outer fold
in Panel B, with fitted models held fixed. $\Delta$ vs.\ clinician is the paired DR difference (positive
favours the policy). Values are estimates of discounted outcome utility under each cohort protocol;
the reward is scaled to $\{-1,+1\}$.}
\label{tab:main}
\centering
\setlength{\tabcolsep}{5pt}
\begin{adjustbox}{max width=\linewidth}
\begin{tabular}{@{}l c c c@{}}
\toprule
Policy & DR value [95\% CI] & FQE-direct [95\% CI]
       & $\Delta$ vs.\ clinician [95\% CI] \\
\midrule
\multicolumn{4}{@{}l}{\textit{Panel A: Held-out MIMIC-III cohort
($n=500$)}} \\
\addlinespace[2pt]
Random              & $-1.498$ [$-1.996$, $+0.838$] & $+0.395$ [$+0.379$, $+0.411$] & $-2.165$ [$-2.664$, $+0.170$] \\
Clinician behaviour & $+0.667$ [$+0.644$, $+0.691$] & $+0.567$ [$+0.553$, $+0.581$] & --- \\
Neural SARSA-Q      & $-0.517$ [$-0.695$, $+1.105$] & $+0.646$ [$+0.639$, $+0.654$] & $-1.183$ [$-1.360$, $+0.437$] \\
Base LLM            & $+0.181$ [$-0.281$, $+1.013$] & $+0.308$ [$+0.290$, $+0.328$] & $-0.485$ [$-0.950$, $+0.341$] \\
\method\ ($N{=}8$, filtered) & \textbf{+0.881} [$+0.762$, $+0.968$] & \textbf{+0.685} [$+0.676$, $+0.695$] & \textbf{+0.214} [$+0.091$, $+0.300$] \\
\midrule
\multicolumn{4}{@{}l}{\textit{Panel B: Independent-site adaptation cohort ($n=571$)}} \\
\addlinespace[2pt]
Random
& $+0.330$ [$+0.249$, $+0.409$]
& $+0.182$ [$+0.163$, $+0.200$]
& $+0.000$ [$-0.063$, $+0.060$] \\
Clinician behaviour
& $+0.330$ [$+0.272$, $+0.389$]
& $+0.211$ [$+0.192$, $+0.228$]
& --- \\
Neural SARSA-Q
& $+0.362$ [$+0.271$, $+0.453$]
& \textbf{+0.267} [$+0.247$, $+0.285$]
& $+0.032$ [$-0.051$, $+0.122$] \\
Base LLM
& $+0.323$ [$+0.248$, $+0.390$]
& $+0.214$ [$+0.195$, $+0.233$]
& $-0.007$ [$-0.069$, $+0.048$] \\
SFT proposal ($N{=}1$)
& $+0.363$ [$+0.255$, $+0.519$]
& $+0.226$ [$+0.207$, $+0.244$]
& $+0.033$ [$-0.055$, $+0.146$] \\
\method\ ($N{=}8$, filtered)
& \textbf{+0.387} [$+0.271$, $+0.544$]
& $+0.244$ [$+0.226$, $+0.261$]
& \textbf{+0.058} [$-0.043$, $+0.172$] \\
\bottomrule
\end{tabular}
\end{adjustbox}
\end{table*}

\subsection{Sources of the estimated gain}

\textbf{Behavioural supervision and value-guided selection.}
Starting from the base LLM, we cumulatively added behavioural
supervision, unfiltered Best-of-8 selection, and the disagreement preference
(Table~\ref{tab:ablation}).
Supervision raised FQE-direct to approximately the clinician
level, whereas DR became unstable. Value-guided selection produced the
largest increase under both estimators. The disagreement preference changed
FQE-direct little, but it brought the DR point estimate back within the
attainable range of discounted utility and narrowed its interval.
The two learning signals play the separate roles
intended by the design. Supervision encourages \emph{behavioural alignment}:
it favours clinician-like proposals but does not guarantee empirical action
coverage or add outcome information. Low trajectory-weight ESS limits the
interpretation of DR at $N=1$: departures from the estimated behaviour model
can compound through per-step ratios over the horizon. A numerical imitation proposal shows the same
pattern (Appendix~\ref{app:controls}). Estimated \emph{value} is added at
selection, which is the only point at which outcome information enters the
policy. The disagreement preference prioritises sampled candidates below
the validation-defined gap threshold; if none passes, selection falls back
to all sampled candidates. The observed interval narrowing describes the
evaluated configurations without establishing calibrated ranking reliability
or an independent source of value.

\begin{table*}[t]
\caption{\textbf{Component ablation with Qwen3-4B on the held-out MIMIC-III cohort.}
Each row cumulatively adds one component. ``$+$ value selection'' uses
unfiltered Best-of-$N$ selection with $N=\Nbon$; ``$+$ disagreement filter''
adds the twin-head disagreement rule in Eq.~\ref{eq:filter}. Both DR and
FQE-direct are reported with 95\% conditional patient-bootstrap intervals as defined in Section~\ref{sec:ope}.}
\label{tab:ablation}
\centering
\setlength{\tabcolsep}{4pt}
\begin{adjustbox}{max width=\linewidth}
\begin{tabular}{@{}l c c@{}}
\toprule
Configuration & DR value [95\% CI] & FQE-direct [95\% CI] \\
\midrule
Base LLM (no SFT) & $+0.181$ [$-0.281$, $+1.013$] & $+0.308$ [$+0.290$, $+0.328$] \\
\quad $+$ SFT (behaviour cloning) & $-0.108$ [$-0.367$, $+1.411$] & $+0.583$ [$+0.570$, $+0.595$] \\
\quad $+$ value selection (Best-of-8, unfiltered) & $+1.022$ [$+0.821$, $+1.195$] & $+0.678$ [$+0.670$, $+0.687$] \\
\quad $+$ disagreement filter (twin-$Q$) & \textbf{+0.881} [$+0.762$, $+0.968$] & \textbf{+0.685} [$+0.676$, $+0.695$] \\
\bottomrule
\end{tabular}
\end{adjustbox}
\end{table*}

\textbf{Number of proposals.}
With the supervised proposal held fixed, we varied the number
of sampled candidates $N$ (Table~\ref{tab:nsweep}).
Most of the gain was obtained by $N=4$, and $N=8$ gave
similar estimates. The number of distinct candidates grew sublinearly with
$N$, while proposal entropy was unchanged by construction.
$N$ controls how strongly selection concentrates
probability on the critic's preferred actions, but only within the
proposal's support (Eq.~\ref{eq:inducedpolicy}). The supervised proposal is
concentrated, with an entropy of about $1.94$ nats, corresponding to roughly
seven effectively used actions. A few draws therefore already cover most of
its high-probability support, and additional draws contribute few new
candidates. Selection can reweight actions within the proposal support but
cannot introduce actions outside that support; this does not ensure
adequate logged-data coverage. We retain $N=8$ as the
primary configuration.

\begin{table*}[t]
\caption{\textbf{Effect of the number of proposals $N$ with Qwen3-4B
(held-out MIMIC-III cohort).} Unique actions are averaged over decision points.
Proposal entropy is the entropy of the fixed proposal distribution and
therefore does not vary with $N$. The $N=1$ row is the SFT proposal without
value selection; the $N=4$ and $N=8$ rows use the disagreement filter.}
\label{tab:nsweep}
\centering
\setlength{\tabcolsep}{3.5pt}
\begin{adjustbox}{max width=\linewidth}
\begin{tabular}{@{}c c c c c@{}}
\toprule
$N$ & DR value [95\% CI] & FQE-direct [95\% CI] & unique actions & proposal entropy \\
\midrule
1 & $-0.108$ [$-0.367$, $+1.411$] & $+0.583$ [$+0.570$, $+0.595$] & $1.00$ & $1.936$ \\
4 & \textbf{+0.977} [$+0.798$, $+1.046$] & \textbf{+0.687} [$+0.677$, $+0.698$] & $3.04$ & $1.936$ \\
8 & $+0.881$ [$+0.762$, $+0.968$] & $+0.685$ [$+0.676$, $+0.695$] & \textbf{4.62} & $1.936$ \\
\bottomrule
\end{tabular}
\end{adjustbox}
\end{table*}

\textbf{Proposal model class and direct use of the critic.}
To test whether the gain depends on the language proposal, we
applied the same selector, critic, and cutoff to a separately trained
numerical behaviour-cloning proposal. To test whether proposal sampling is
needed at all, we also used the selection critic directly as a greedy or
Boltzmann policy over all 25 actions (Appendix~\ref{app:controls}).
Before selection, the numerical and language proposals had
the same FQE-direct estimate, and value-guided selection increased
FQE-direct for both. Greedy maximisation of the critic reached a similar
FQE-direct estimate to proposal-constrained selection, but the cap clipped
most of its stepwise importance ratios ($91.8\%$, compared with $47$--$48\%$
for Best-of-8). The Boltzmann critic policy had a lower estimate.
The estimated gain is a property of the selection
operator applied to a behaviour-supervised proposal, not of the language
model class. The role of the language model in \method\ is therefore
structural: the interface turns an LLM recommender into a policy to which
the operator applies and that OPE can evaluate. The direct-critic controls
show why the proposal step is retained. Proposal-constrained selection
reaches comparable FQE-direct estimates with fewer clipped stepwise
importance ratios than direct critic maximisation. This is an empirical
overlap diagnostic, rather than a guarantee of sufficient logged-data
coverage. For the numerical proposal, the disagreement
preference again lowered FQE-direct relative to unfiltered selection,
consistent with its role as a conservative option.

\textbf{Backbone scale.}
We repeated the action-policy pipeline with a \qwenbig\
proposal while sharing the selection critic, clinician reference, and OPE
protocol (Table~\ref{tab:scale}).
The larger base model was much closer to the clinician
reference than the smaller one. After supervision and selection, however,
both backbones reached similar estimated value and a similar paired DR
difference from the clinician reference ($+0.226$, 95\% interval
$[+0.082,+0.323]$, at 14B). At 14B, the disagreement preference markedly
narrowed the DR interval relative to unfiltered selection.
A larger backbone carries a stronger zero-shot prior
over treatments. Once the proposal is behaviour-supervised, however, the
value of the selected policy is governed mainly by the shared critic and
selection operator, so the estimated gain does not depend on a particular
backbone size. The narrower 14B interval with the disagreement preference
illustrates its stabilising role when the proposal assigns mass to
candidates with greater twin-head disagreement.

\begin{table*}[t]
\caption{\textbf{Within-family backbone-scale analysis: Qwen3-4B versus Qwen3-14B.}
Both backbones use the same held-out 500-patient cohort, action-policy
definition, selection critic, and OPE protocol. The clinician row is a shared
backbone-independent reference. Intervals use the conditional patient-bootstrap protocol of Section~\ref{sec:ope}.
Proposal training settings are given in Appendix~\ref{app:hparams}.}
\label{tab:scale}
\centering
\setlength{\tabcolsep}{3pt}
\begin{adjustbox}{max width=\linewidth}
\begin{tabular}{@{}l c c c c@{}}
\toprule
Policy & 4B DR [95\% CI] & 4B FQE-direct & 14B DR [95\% CI] & 14B FQE-direct [95\% CI] \\
\midrule
Base LLM & $+0.181$ [$-0.281$, $+1.013$] & $+0.308$ & $+0.665$ [$+0.501$, $+0.954$] & $+0.512$ [$+0.498$, $+0.528$] \\
SFT only ($N{=}1$) & $-0.108$ [$-0.367$, $+1.411$] & $+0.583$ & $-0.425$ [$-1.259$, $+0.436$] & $+0.534$ [$+0.521$, $+0.546$] \\
\method\ ($N{=}4$) & \textbf{+0.977} [$+0.798$, $+1.046$] & \textbf{+0.687} & \textbf{+0.918} [$+0.631$, $+0.977$] & $+0.653$ [$+0.643$, $+0.664$] \\
\method\ ($N{=}8$) & $+0.881$ [$+0.762$, $+0.968$] & $+0.685$ & $+0.892$ [$+0.760$, $+0.985$] & \textbf{+0.722} [$+0.713$, $+0.732$] \\
Unfiltered \method\ ($N{=}8$) & $+1.022$ [$+0.821$, $+1.195$] & $+0.678$ & $+0.857$ [$+0.099$, $+1.081$] & $+0.699$ [$+0.691$, $+0.708$] \\
Clinician behaviour (shared) & $+0.667$ [$+0.644$, $+0.691$] & $+0.567$ & $+0.667$ [$+0.644$, $+0.691$] & $+0.567$ [$+0.553$, $+0.581$] \\
\bottomrule
\end{tabular}
\end{adjustbox}
\end{table*}

\subsection{Reuse of a frozen language proposal at an independent site}
The MIMIC-III proposal was applied unchanged to a second ICU cohort with
different case mix and higher mortality, while the critic, behaviour model,
and evaluation models were fitted locally by cross-fitting
(Table~\ref{tab:main}, Panel B).
At the independent site, the supervised proposal had higher estimated value
than the base LLM, and local selection
increased it further, under both estimators. \method\ had the highest DR
estimate among the evaluated policies, whereas Neural SARSA-Q had the
highest FQE-direct estimate. The interval for the paired DR difference from
the clinician reference included zero.
The frozen language proposal retained source-domain input processing,
while numerical components were fitted locally. Both DR importance ratios
and FQE policy expectations used the overlap-constrained target
probabilities. This analysis evaluates proposal reuse under local numerical
adaptation.

\subsection{Inspecting the explicit policy and its rationale}

\begin{figure}
  \centering
  \includegraphics[width=0.95\linewidth]{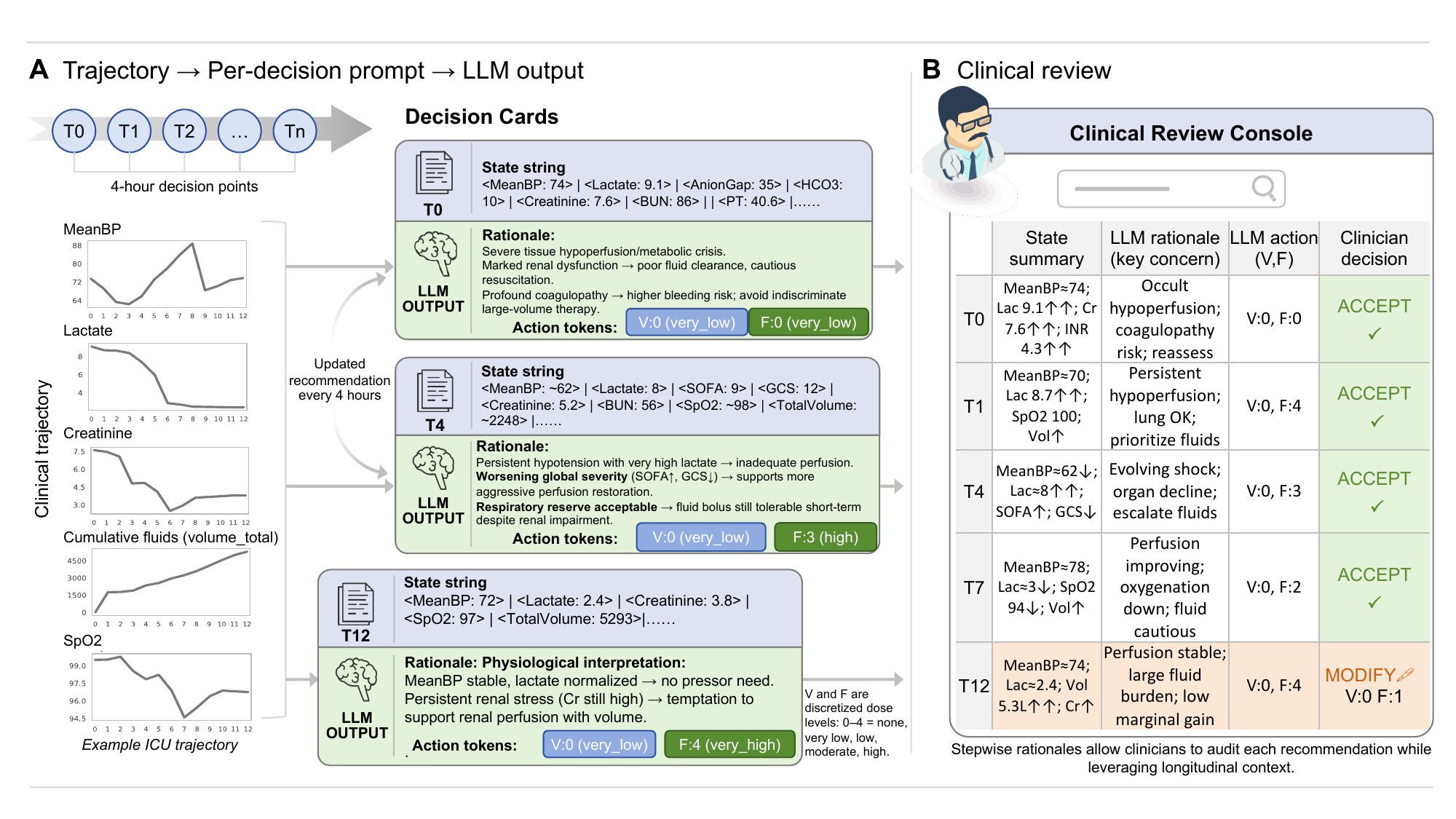}
  \caption{\textbf{Illustrative decision interface and review sequence.}
  \textbf{a}, Each state is converted into a selected action and a rationale
  generated after selection. \textbf{b}, Schematic review: at $T_{12}$ the
  reviewer lowers the fluid level from $F{:}4$ to $F{:}1$ given stable MeanBP,
  normalised lactate, and cumulative fluid burden. The review entries
  illustrate the interface; the evaluated action policy does not include these
  edits. Rationales are action-conditioned descriptions.}
  \label{fig:example}
\end{figure}

\textbf{Illustrative decision interface.}
Figure~\ref{fig:example} shows how a clinician-facing review could use the
outputs: each 4-hour state is paired with the selected action and an
action-conditioned rationale. At $T_{12}$, the displayed recommendation
continues high-volume fluid despite stable MeanBP, normalised lactate, and
substantial cumulative fluid, and a schematic reviewer lowers the fluid
level. Because the observation, the action, and the rationale are displayed
together, such a discrepancy is visible at the point of review. The edit is
an interface demonstration and is not part of the evaluated policy.

Because \method\ defines exact action probabilities and generates its
rationale separately, its behaviour can be examined across the population,
rather than through individual sampled outputs.

\begin{figure}
\centering
\includegraphics[width=0.90\linewidth]{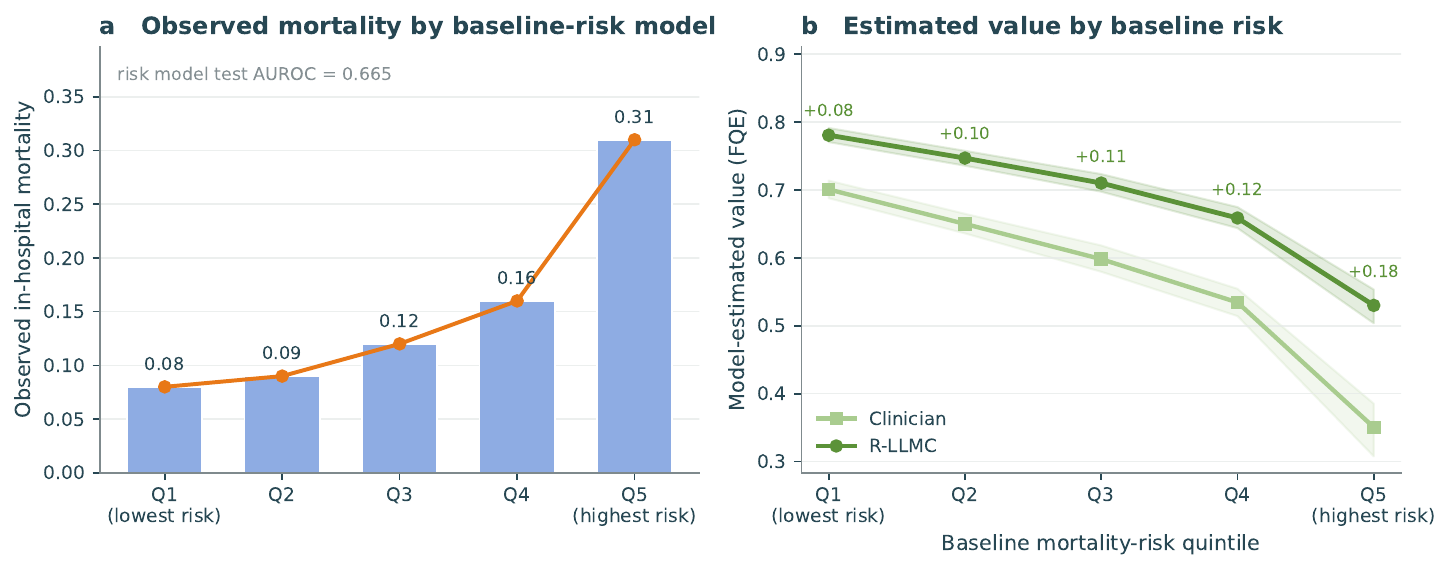}
\caption{\textbf{Baseline-risk stratification and model-estimated policy
value.} \textbf{a}, Observed in-hospital mortality across test-set quintiles
defined by an independent baseline-risk model. \textbf{b}, Mean FQE-estimated
initial-state value for clinician behaviour and R-LLMC within the same strata;
bands show patient-bootstrap 95\% confidence intervals and annotations give
the difference. Panel-b values are model estimates, not death probabilities.}
\label{fig:mortality}
\end{figure}

\textbf{Estimated value across baseline risk.}
An independent logistic model fitted on training data divided
test patients into quintiles of predicted baseline risk, across which
observed mortality increased monotonically (Fig.~\ref{fig:mortality}a). We
compared the mean FQE-estimated initial-state value of \method\ and the
clinician reference within each quintile.
\method\ had a higher estimated value in every quintile, and
the difference was largest in the highest-risk quintile
(Fig.~\ref{fig:mortality}b).
These are FQE-estimated utilities, not mortality probabilities.

\begin{figure}
\centering
\includegraphics[width=0.9\linewidth]{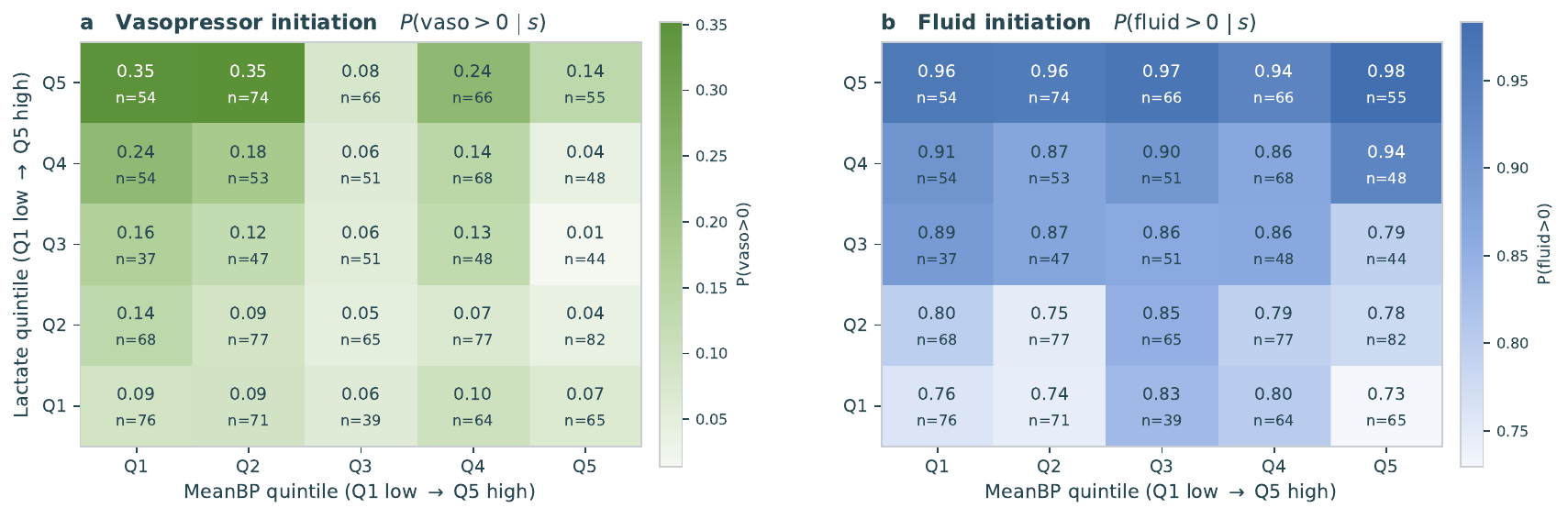}
\caption{\textbf{Early-resuscitation treatment landscape.}
\textbf{a}, Vasopressor-initiation probability and \textbf{b},
fluid-initiation probability under R-LLMC, stratified jointly by MeanBP
and lactate quintile at the first three decision points. Each cell reports
the policy probability and number of contributing states; colour scales
are panel-specific. The $1{,}500$ states comprise three observations per
test patient.}
\label{fig:heatmap}
\end{figure}

\textbf{Treatment allocation across physiology.}
We computed the exact probabilities of vasopressor and fluid
initiation under \method\ at the first three decision points of each test
trajectory and mapped them over joint MeanBP and lactate quintiles
(Fig.~\ref{fig:heatmap}).
Vasopressor initiation was concentrated in the hypotensive,
hyperlactataemic region and was rare in stable regions. Fluid initiation was
common throughout and increased with lactate. Adjusted patient-cluster
bootstrap estimates confirmed that vasopressor initiation increased with
lower MeanBP and higher lactate, and that fluid initiation increased with
lactate.
The selected policy allocates treatment along
recognisable haemodynamic gradients. These maps are read directly from the
policy's action probabilities rather than estimated from sampled
recommendations, which is a direct consequence of the explicit distribution
and cannot be obtained in this way from a free-text recommender. The
associations are descriptive and are not evaluated against guidelines.

\begin{figure}
\centering
\includegraphics[width=0.75\linewidth]{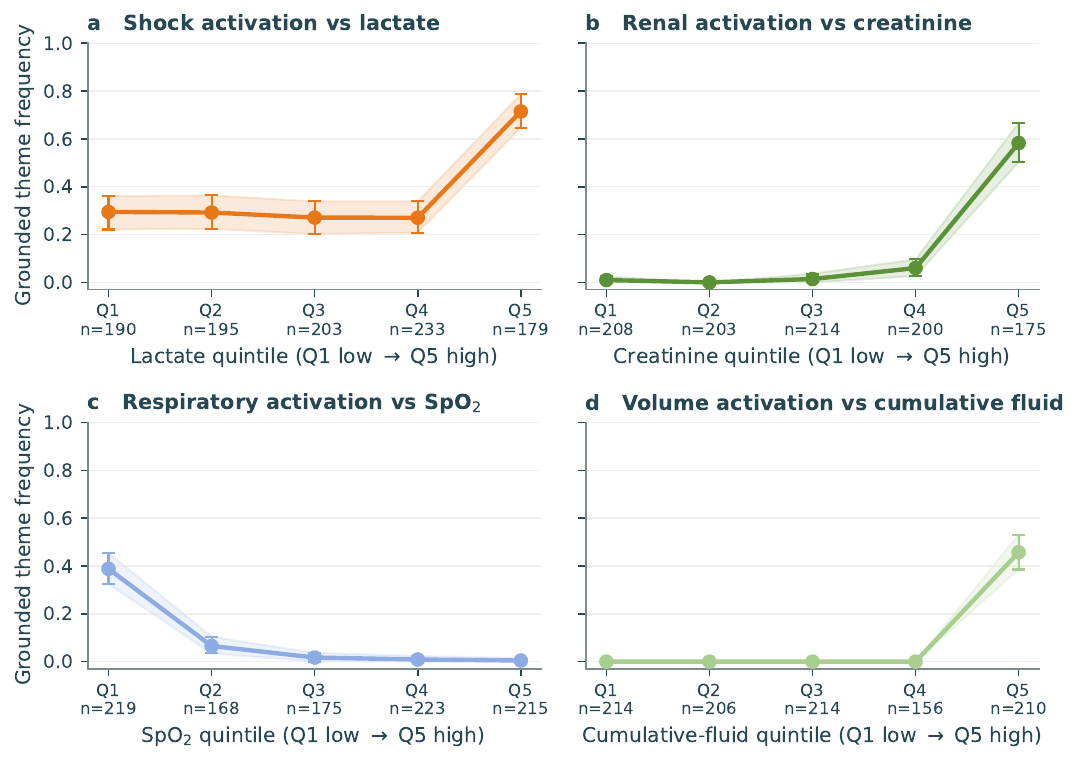}
\caption{\textbf{Physiology--theme associations in fixed-action rationales.}
With the conditioning action fixed for each decision, grounded theme
frequency is compared across quintiles of the corresponding
physiological variable: \textbf{a}, shock versus lactate; \textbf{b}, renal
dysfunction versus creatinine; \textbf{c}, respiratory compromise versus
SpO$_2$; \textbf{d}, volume status versus cumulative fluid. Points are
observed frequencies, error bars are patient-cluster bootstrap 95\% confidence
intervals, and the number below each quintile is the record count
($n=1{,}000$ decisions from 500 test patients).}
\label{fig:grounding}
\end{figure}

\textbf{Evidence associations of action-conditioned rationales.}
For each of 1,000 test decisions, the rationale model received
a fixed action and the available past-only observations
(Appendix~\ref{app:language}). A clinical theme was counted as active only
when the rationale both declared it and cited an abnormal-direction value of
a pre-specified theme-specific variable. Activation was then compared across
quintiles of that variable (Fig.~\ref{fig:grounding}).
For shock, renal dysfunction, volume status, and respiratory
compromise, activation rose toward the clinically abnormal end of the
corresponding variable. Coagulation could not be estimated, because grounded
declarations were too rare.
These associations relate rationale themes to visible
physiological evidence under fixed-action conditioning. The model receives
no critic values, and the analysis does not reconstruct the computation used
for action selection.

\section{Discussion}

\method\ addresses three requirements that existing approaches meet only in
part: a recommendation should specify the next treatment and relate it to
delayed outcome, it should be informed by retrospective treatment data and
be evaluable from those records, and the
clinical context behind it should be open to review. Numerical RL policies
meet the first and are evaluable, but expose only a state vector and a dose
index. LLM recommenders expose clinical context, but define neither an
explicit treatment distribution nor a link to outcome. The framework meets
all three by assigning each requirement to a separate component connected
through a shared action vocabulary. The language proposal defines the
treatment distribution and its support. The numerical critic introduces
outcome information through selection within that support. The closed-form
induced policy makes the result evaluable, and the post-selection rationale
exposes the displayed evidence without influencing the action.

The experiments indicate how these components divide the work. Behavioural
supervision encourages behavioural alignment: it favours clinician-like
proposals but does not guarantee adequate logged-data coverage or add outcome
information. Estimated value is added at
selection, the only point at which outcome information enters the policy.
The same pattern held when the language proposal was replaced by a
numerical behaviour-cloning proposal, and at both backbone sizes. The
estimated gain is therefore a property of the selection operator applied to
a behaviour-supervised proposal rather than of the language model class or
its scale. Two further results show why selection is constrained to the
proposal. Gains saturated with the number of candidates, while selection
remained within the proposal support. Direct maximisation of the critic over
all actions reached a similar FQE-direct estimate but required more clipped
importance ratios under the reported OPE protocol.
The disagreement preference acted as a conservative mechanism that
stabilised evaluation rather than as an independent source of value.

These results define the role of the language model in the framework. The
evidence does not support a claim that a language proposal yields higher
estimated value than a numerical one. Its contribution is structural: the
interface turns an LLM recommender into an explicit policy that offline
value information can improve and OPE can evaluate on the same scale as
numerical policies, while its inputs and outputs remain in clinical terms.
The base LLM illustrates this. Used directly, it was a poorly performing
policy, and the interface made that deficit measurable rather than hidden
behind fluent output. Because the selector and evaluation depend only on
proposal probabilities, the proposal can be replaced without changing either
of them. This modularity also allowed a frozen language proposal to be
reused at an independent site while the numerical components were refitted
locally. The site estimates were smaller and their clinician comparison
included zero, and a frozen numerical proposal could be reused in the same
way. The analysis therefore establishes that such reuse is feasible, not
that it confers a language-specific transfer advantage.

The rationale model is deliberately separated from the decision. Generating
text only after the action is fixed means that an explanation, however
fluent, cannot alter the treatment. Accordingly, its theme declarations
tracked the displayed physiology, while treatment allocation responded to
the same inputs through the proposal and the critic. The rationale should
therefore be read as an action-conditioned description of the visible
evidence, not as a reconstruction of the numerical computation that selected
the action, a distinction emphasised in work on rationale faithfulness and
explanation
standards~\citep{turpin2023language,arrieta2020explainable,longo2024explainable}.
The explicit policy distribution complements this case-level description at
the population level: treatment allocation can be read directly from action
probabilities across observed clinical states, as in the resuscitation
landscape, rather than inferred from sampled outputs.

\textbf{Limitations.}
All value estimates are retrospective. Exact target probabilities make OPE
applicable, but they do not make it precise. Trajectory-weight ESS was low
for every policy that departed from the estimated behaviour model, including
an imitation policy, and a large fraction of importance ratios was clipped.
The bootstrap intervals are conditional on fitted models and do not include
retraining variability. Unmeasured clinical factors may also affect the
estimates. Reported values are therefore estimates of discounted outcome
utility under the specified protocol, not identified treatment effects or
changes in mortality. The action space is a discretised $5\times5$
vasopressor--fluid grid, reduced to five vasopressor levels at the
independent site, and the terminal reward does not encode intermediate
physiological targets. The language proposal uses only the current
structured observation. It therefore does not yet exploit clinical notes or
longer histories, the settings in which a language interface may contribute
most beyond a numerical one. Results are reported for one model family and
two cohorts. Rationale targets were derived from model-generated
explanations of logged actions rather than authored by clinicians, and the
illustrative review sequence demonstrates the interface without evaluating
clinician use. Prospective evaluation would be required before any clinical
application.

\section{Conclusion}

We presented \method, a framework that makes an LLM treatment recommender
an explicit and evaluable policy for sepsis resuscitation. Constrained
decoding turns the language proposal into an exact distribution over a
clinical action vocabulary. A frozen numerical critic introduces
delayed-outcome information by selecting among proposal samples, without
outcome-derived gradients to the proposal model and without leaving the
proposal's support. The induced policy has a closed form that supplies the
target probabilities required for off-policy evaluation, and a separate
model explains the selected action without influencing it. On held-out
MIMIC-III patients, value-guided selection improved estimated outcome value
over both the supervised proposal and an estimated clinician reference.
Controls attributed this gain to the selection operator rather than to the
proposal model class, and a frozen language proposal could be reused at an
independent site with locally fitted numerical components. The framework
provides a basis for developing language-based treatment policies that can be
improved with outcome data and evaluated on the same terms as conventional
treatment policies, with prospective evaluation as the necessary next step.

\appendix

\section{Trajectory construction and preprocessing}
\label{app:data}

Each MIMIC-III trajectory is represented on a 4-hour decision grid using the
following 20 physiological and treatment-context variables:
\texttt{HeartRate}, \texttt{SysBP}, \texttt{DiasBP}, \texttt{MeanBP},
\texttt{RespRate}, \texttt{SpO2}, \texttt{SOFA\_24hours}, \texttt{GCS\_min},
\texttt{ANIONGAP}, \texttt{BICARBONATE}, \texttt{CREATININE},
\texttt{CHLORIDE}, \texttt{LACTATE}, \texttt{PLATELET}, \texttt{INR},
\texttt{PT}, \texttt{BUN}, \texttt{GLUCOSE}, \texttt{max\_dose\_vaso}, and
\texttt{volum\_total}.

Within each 4-hour interval, the exported vasopressor series is aggregated
by maximum and fluid amounts by summation. Their cumulative maximum and sum,
respectively, form the treatment-context variables.

Past-only forward filling imputes missing measurements, followed by
training-partition medians for unresolved entries. Training-partition feature
means and standard deviations are held fixed thereafter.

Actions follow Eq.~\ref{eq:action}, using exposure in the next 4-hour interval
as the target for state row $j$; state variables stop at $j$, and the final
state has no outgoing action.
Cut points fitted to positive training exposures were fixed for validation
and test. On the preprocessed scale, the vasopressor cuts were approximately
$(0.120000,0.325142,1.039423)$, and the fluid cuts were
$(65.212,292.750,626.099)$, rounded for reporting.
For cut points $q_1,q_2,q_3$, positive exposures occupy bins
$(0,q_1]$, $(q_1,q_2]$, $(q_2,q_3]$, and $(q_3,\infty)$; zero exposure forms
bin 0.

\section{Language-interface implementation}
\label{app:language}

The proposal is supervised to emit action-only completions of the form
\texttt{V:<v> F:<f>}. Each conditional distribution is obtained by restricting
logits to valid digit tokens $0,\ldots,4$, dividing by $T_{\mathrm{s}}=0.7$,
and applying softmax. Vasopressor probabilities and the five
vasopressor-conditioned fluid distributions are combined as in
Eq.~\ref{eq:parsedpolicy}. Candidates are sampled independently with
replacement from this joint distribution. In the internal analysis, these
materialised proposal probabilities define the selector and its induced
policy used for OPE. The proposal uses the current observation
only ($H_{\mathrm{prop}}=1$).

The proposal and rationale modules are fitted separately with completion-only
supervision. Neither objective includes an outcome-based loss.
Proposal supervision contributes to the action policy evaluated by OPE;
the rationale module describes the fixed action in relation to displayed
observations. The teacher-generated source explanations were produced with
access to full retrospective trajectories and outcomes. Rationale targets were reconstructed from the
archived observed-action explanations after alignment with logged actions;
outcome and counterfactual fields were excluded from student supervision.
Following lexical screening, feature statements were checked against
past-only histories, with source themes constraining selection, and matched
evidence was rendered anew into the fixed \texttt{themes}, \texttt{evidence},
and \texttt{rationale} schema. Evidence entries identify the feature,
relative observation time, value, and qualitative direction. The corpus
comprised 2,464 training and 279 validation examples. The rationale module
conditions on up to four past-only observations and a fixed action---logged
during training and selected at inference---without selection-critic values,
disagreement scores, or terminal outcomes. Its parameters and outputs do not
participate in action selection or OPE. Removing explicit outcome fields
and reconstructing targets from past-only evidence cannot rule out residual
hindsight bias inherited from the teacher-generated explanations. This
limitation concerns rationale supervision and its descriptive evaluation;
the rationale model does not participate in action selection or OPE.

For the deterministic rationale analysis, each decision's conditioning
action is held fixed while theme declarations are compared with the visible
past-only observations. This analysis concerns action-conditioned text;
OPE evaluates the stochastic Best-of-$N$ action distribution separately.

\section{Off-policy evaluation implementation}
\label{app:ope}

\subsection{Policy and estimator definitions}
The following estimator definitions describe the internal analysis. Let
$\pi$ denote the target action distribution: \method\ uses
Eq.~\ref{eq:inducedpolicy}, and the other policies use their corresponding
25-action distributions. At the independent site, joint proposals are first
marginalised over fluid to five vasopressor levels before selection and OPE.
Each held-out fold uses a locally fitted five-action behaviour policy and
overlap-constrained target probabilities; these same probabilities enter
both DR importance ratios and FQE policy expectations.
The behaviour policy $\mu(a\mid x)$ is fitted from logged clinician actions on
the training partition and selected using the validation partition. Small
probabilities are floored at $10^{-4}$ and the distribution is renormalised.
No test outcomes are used in fitting the behaviour policy.

A separate fitted Q evaluation (FQE) model is trained for each target policy.
These models are fitted separately from the selection critic used by R-LLMC.
Three independently initialised FQE models are trained for each policy. Their
predictions are averaged:
\begin{equation}
\widehat Q^\pi(x,a)
=
\frac{1}{3}
\sum_{m=1}^{3}
\widehat Q^\pi_m(x,a),
\end{equation}
and
\begin{equation}
\widehat V^\pi(x,o)
=
\sum_{a\in\mathcal A}
\pi(a\mid x,o)\widehat Q^\pi(x,a).
\label{eq:app_fqe_value}
\end{equation}

DR and FQE-direct use the same averaged FQE predictions. The
complementary FQE-direct estimate is
\begin{equation}
\widehat V^\pi_{\mathrm{FQE}}
=
\frac{1}{n}
\sum_{i=1}^{n}
\widehat V^\pi(x_{i,0},o_{i,0}).
\label{eq:app_fqe_direct}
\end{equation}

For patient trajectory $i$ and decision time $t$, the clipped log importance
ratio is
\begin{equation}
\ell_{i,t}
=
\operatorname{clip}
\left(
\log
\frac{
\pi(a_{i,t}\mid x_{i,t},o_{i,t})
}{
\mu(a_{i,t}\mid x_{i,t})
},
-\log20,
+\log20
\right).
\end{equation}
The cumulative and self-normalised weights are
\begin{equation}
W_{i,t}
=
\exp
\left(
\sum_{k=0}^{t}\ell_{i,k}
\right),
\qquad
\widetilde W_{i,t}
=
\frac{W_{i,t}}
{\sum_{j=1}^{n}W_{j,t}}.
\label{eq:app_weights}
\end{equation}

The primary estimator is
\begin{equation}
\adjustbox{max width=0.94\linewidth}{$\displaystyle
\widehat V_{\mathrm{DR}}^\pi
=\frac{1}{n}\sum_{i=1}^{n}\widehat V^\pi(x_{i,0},o_{i,0})
+\sum_{t=0}^{T_{\max}-1}\gamma^t\sum_{i=1}^{n}\widetilde W_{i,t}
\Big[r_{i,t}+\gamma(1-d_{i,t})\widehat V^\pi(x_{i,t+1},o_{i,t+1})
-\widehat Q^\pi(x_{i,t},a_{i,t})\Big].$}
\label{eq:app_dr}
\end{equation}

Following the final observed transition, absorbing-state padding sets
subsequent temporal-difference residuals to zero and carries the last
cumulative weight forward, equivalent to unit importance ratios during
padding. All trajectories remain in each time-step normalisation denominator.

The unclipped sensitivity analysis uses the same estimator but replaces
$\ell_{i,t}$ by the untruncated log importance ratio.

\subsection{Independent-site overlap constraint}
\label{app:site_overlap}

At the independent site, joint language and numerical proposals are first
marginalised over fluid, and any Best-of-$N$ selection then operates over
five vasopressor levels. For outer fold $k$, let $q_k(a\mid x)$ denote the
resulting raw target policy, or the corresponding five-action baseline, and
let $\mu_k(a\mid x)$ denote the local behaviour model. The latter uses
softmax temperature $1$, followed by the $10^{-4}$ probability floor and
renormalisation described above. The evaluated target is a convex mixture
with this fitted behaviour reference:
\begin{equation}
\pi_k^{\mathrm{site}}(a\mid x)
=\mu_k(a\mid x)+\lambda_k^{\mathrm{eff}}(x)
\bigl[q_k(a\mid x)-\mu_k(a\mid x)\bigr].
\label{eq:site_mixture}
\end{equation}
The fold-level coefficient $\lambda_k\in[0,1]$ is reduced at each state when
needed to bound the action-probability ratios. Define
\begin{equation}
r_k(a,x)=\frac{q_k(a\mid x)}{\mu_k(a\mid x)},
\qquad
r_k^{\max}(x)=\max_a r_k(a,x),
\qquad
r_k^{\min}(x)=\min_a r_k(a,x).
\end{equation}
With $C=20$ and the numerical boundary $C'=C(1-10^{-12})$, set
\begin{equation}
\begin{aligned}
U_k(x)&=
\begin{cases}
\dfrac{C'-1}{r_k^{\max}(x)-1},&r_k^{\max}(x)>1,\\
1,&\text{otherwise},
\end{cases}
&
L_k(x)&=
\begin{cases}
\dfrac{1-1/C'}{1-r_k^{\min}(x)},&r_k^{\min}(x)<1,\\
1,&\text{otherwise},
\end{cases}\\
\lambda_k^{\mathrm{eff}}(x)&=
\min\{\lambda_k,U_k(x),L_k(x)\}.
\end{aligned}
\label{eq:site_ratio_cap}
\end{equation}
Thus $1/C\leq\pi_k^{\mathrm{site}}(a\mid x)/\mu_k(a\mid x)\leq C$
for every action. The statewise reduction uses only $q_k$ and $\mu_k$, not
the logged action, reward or outcome. Raw target probabilities receive no
additional floor; the final mixture is checked for non-negativity and
normalisation. This constraint bounds ratios relative to a fitted behaviour
model, rather than guaranteeing adequate empirical action coverage. The
mixture can assign mass outside the raw target support; proposal-support
preservation applies to the Best-of-$N$ operator before this site-specific
transformation.

Within each outer-training partition, $15\%$ of patients form an
outcome-stratified inner-validation split. Local behaviour models are fitted
on the inner-training split and selected by inner-validation action
cross-entropy. Each of the five vasopressor levels must have at least five
logged actions in the inner-training split before overlap calibration.
The ratio cap, probability floor, following candidate grid and diagnostic
thresholds are fixed; only $\lambda_k$ is selected within the fold:
\begin{equation}
\Lambda=\{0,0.025,0.05,0.10,0.20,0.40,0.60,0.80,1\}.
\end{equation}
For each candidate, cumulative importance weights are computed on
inner-validation trajectories using the transformed target probabilities.
After termination, the last cumulative weight is carried forward. If
$u_{i,t}$ denotes the normalised weight at time $t$, then
$\mathrm{ESS}_t=1/\sum_i u_{i,t}^2$. For $n_{\mathrm{val}}$ validation
patients, every time point must satisfy
$\mathrm{ESS}_t\geq\min\{n_{\mathrm{val}},\max(10,0.10n_{\mathrm{val}})\}$
and $\max_i u_{i,t}\leq\max\{0.10,1/n_{\mathrm{val}}\}$, with zero
stepwise ratio clipping. The largest candidate meeting all conditions
simultaneously for the Base LLM, SFT proposal, filtered Best-of-4,
filtered Best-of-8 and unfiltered Best-of-8 policies is selected.
The $\lambda_k=0$ candidate returns the fitted behaviour policy; calibration
fails if no candidate passes. The same fold-level coefficient is then
applied to every non-clinician target policy, including Random, Neural
SARSA-Q and the numerical-proposal controls, which do not influence its
selection. Effective statewise coefficients may differ between policies.
The calibration criteria use states, logged actions and fitted probabilities,
without reward, outcome, DR or FQE values as selection objectives. Both FQE
policy expectations and DR importance ratios use
$\pi_k^{\mathrm{site}}$ from Eq.~\ref{eq:site_mixture}; the fitted clinician
reference remains $\mu_k$.

\subsection{Resampling and implementation}
Each FQE ensemble is fitted once on the corresponding development partition.
Percentile 95\% patient-bootstrap intervals use $B=500$ trajectory resamples
stratified by terminal outcome. Policy differences from the clinician reference
use matched bootstrap indices. All fitted models are held fixed. For the
independent-site analysis, pooled out-of-fold predictions are resampled
within joint outer-fold and terminal-outcome strata; models are not refitted
or cross-fitted again within bootstrap replicates.

\subsection{Evaluation diagnostics}
Trajectory diagnostics use the final cumulative weight $W_i^{\mathrm{end}}$.
Let $u_i=W_i^{\mathrm{end}}/\sum_jW_j^{\mathrm{end}}$ and $w_i=nu_i$.
The $u_i$ sum to one, whereas the reported diagnostic weights $w_i$ have mean
one. Effective sample size is
$\mathrm{ESS}=1/\sum_i u_i^2=n^2/\sum_iw_i^2$.
Table~\ref{tab:diag} reports these trajectory-weight and per-step clipping
diagnostics for all internal policies. When the target equals the estimated
behaviour model, ratios are one, giving ESS $=500$ and maximum weight $=1$.
This is an implementation identity, rather than a validation of the fitted
behaviour model.

\begin{table*}[t]
\caption{\textbf{Off-policy evaluation diagnostics on the held-out
cohort.} ESS\% is trajectory-level effective sample size as a percentage
of $n=500$. Clip\% is the fraction of stepwise log importance ratios
affected by the pre-specified $\log(20)$ cap. Maximum weight is
$\max_iw_i$ on the mean-one scale; extreme weights satisfy $w_i>10$.}
\label{tab:diag}
\centering
\begin{adjustbox}{max width=\linewidth}
\begin{tabular}{@{}l c c c c c@{}}
\toprule
Policy & ESS & ESS\% & clip\% & max weight & extreme weights (\%) \\
\midrule
Random              & $1.0$ & $0.2$ & $0.84$  & $494.2$ & $0.20$ \\
Clinician behaviour & 500.0 & 100.0 & 0.00 & 1.0 & 0.00 \\
Neural SARSA-Q      & $1.0$ & $0.2$ & $17.63$ & $499.9$ & $0.20$ \\
Base LLM            & $2.1$ & $0.4$ & $88.89$ & $316.7$ & $1.00$ \\
\method\ ($N{=}8$) & $1.9$ & $0.4$ & $47.06$ & $333.4$ & $0.60$ \\
\bottomrule
\end{tabular}
\end{adjustbox}
\end{table*}

For filtered \method\ at $N=8$, the DR point estimate was $0.881$ with
per-step log ratios capped at $\pm\log20$ and $0.877$ without that cap.
These two settings gave similar point estimates. With a single terminal
reward in $\{-1,+1\}$ and $\gamma\leq1$, true discounted returns lie in
$[-1,1]$, whereas finite-sample DR estimates need not remain within that range.
The reported Random and unfiltered-selection estimates are retained on their
original scale.

\subsection{Interpretation of retrospective estimates}
Sequential clinical OPE depends on modelling and data coverage because logged
trajectories contain outcomes for administered treatments. Policy differences
can concentrate importance weights on a small portion of the cohort, limiting
the information available to weighted
estimators~\citep{thomas2016data,komorowski2018artificial}. Here the
trajectory-weight ESS for \method\ was $1.9/500$, and $47.06\%$ of stepwise
ratios were clipped.
Low trajectory ESS is not specific to value-guided selection: a numerical
proposal fitted to clinician actions had a trajectory ESS of $2.46/500$
(Appendix~\ref{app:controls}).
Lee et al.~\citep{lee2024reinforcement} reported a WIS
ESS of $3.33/2531$ ICU admissions ($0.13\%$) and related it to differences
between learned and clinician policies. This is context for a shared
evaluation difficulty, not a reliability threshold for the present estimator.
Trajectory-weight ESS is distinct from the sample size used to fit FQE.
FQE depends on function approximation and data coverage, and DR and FQE-direct
share its predictions. The bootstrap intervals condition on fitted models and
outcome-stratified resampling; they do not incorporate model-retraining
uncertainty. These features, together with possible unmeasured clinical factors,
bound the retrospective interpretation: reported values estimate discounted
outcome utility under the specified protocol, rather than identified treatment
effects or changes in mortality.

\section{Numerical-proposal and direct-critic controls}
\label{app:controls}

These controls examine two questions: whether the selection interface
requires a language proposal, and how proposal-constrained selection compares
with using the selection critic directly as a policy. Internal controls use
the patient partitions, preprocessing, frozen selection critic, disagreement
cutoff $\xi$, and OPE protocol of Section~\ref{sec:ope} and
Appendix~\ref{app:ope}. Each target policy has its own three-member FQE
ensemble. Paired differences are computed among the control policies using
shared outcome-stratified bootstrap indices ($B=500$), with fitted models
held fixed.

\subsection{Internal control policies}

The \textbf{MLP proposal} ($N=1$) is a 20--256--256--25 LayerNorm/ReLU
network trained by cross-entropy on logged training actions, with early
stopping on validation cross-entropy and proposal temperature
$T_{\mathrm{s}}=0.7$. It shares the architecture of the
clinician-behaviour model but is trained separately from a different
initialisation; its weights do not define the OPE behaviour reference.
\textbf{MLP Best-of-8} uses Eq.~\ref{eq:inducedpolicy} with the MLP proposal
in place of $p_t$ and $N=8$, with or without the disagreement preference.
\textbf{Critic greedy} selects
$\arg\max_{a\in\mathcal A}\bar Q_\omega(x_t,a)$ over all 25 actions,
breaking ties by the smaller action index. \textbf{Critic softmax} uses
$\pi(a\mid x_t)\propto\exp\!\big(\bar Q_\omega(x_t,a)/\kappa\big)$ over all
25 actions, with $\kappa$ selected on validation states to match the
clinician-behaviour model's mean action entropy; no test outcomes or OPE
estimates were used.

\subsection{Internal cohort}

\begin{table}[t]
\caption{\textbf{Numerical-proposal and direct-critic controls on the
held-out MIMIC-III cohort ($n=500$).} DR uses the per-step log-ratio cap of
$\log 20$. ESS is the trajectory-level effective sample size of
Appendix~\ref{app:ope}; clip\% is the fraction of stepwise log importance
ratios affected by the cap. Intervals are 95\% conditional patient-bootstrap
intervals.}
\label{tab:ctrl_internal}
\centering
\setlength{\tabcolsep}{4pt}
\begin{adjustbox}{max width=\linewidth}
\begin{tabular}{@{}l c c c c@{}}
\toprule
Policy & DR [95\% CI] & FQE-direct [95\% CI] & ESS & clip\% \\
\midrule
MLP proposal ($N{=}1$)
  & $-0.481$ [$-0.819$, $+0.896$] & $+0.583$ [$+0.570$, $+0.597$] & $2.46$ & $0.19$ \\
MLP Best-of-8, filtered
  & $+0.880$ [$+0.788$, $+0.999$] & $+0.711$ [$+0.701$, $+0.722$] & $1.46$ & $48.34$ \\
MLP Best-of-8, unfiltered
  & $+0.887$ [$+0.760$, $+1.002$] & $+0.769$ [$+0.760$, $+0.777$] & $1.93$ & $47.71$ \\
Critic greedy
  & $+0.931$ [$+0.847$, $+1.057$] & $+0.684$ [$+0.675$, $+0.694$] & $1.09$ & $91.79$ \\
Critic softmax
  & $+1.136$ [$+0.858$, $+1.341$] & $+0.617$ [$+0.609$, $+0.626$] & $2.71$ & $18.11$ \\
\bottomrule
\end{tabular}
\end{adjustbox}
\end{table}

\begin{table}[t]
\caption{\textbf{Paired differences among internal control policies.}
Differences are policy minus the MLP proposal at $N=1$, computed with shared
bootstrap indices.}
\label{tab:ctrl_internal_paired}
\centering
\setlength{\tabcolsep}{4pt}
\begin{adjustbox}{max width=\linewidth}
\begin{tabular}{@{}l c c@{}}
\toprule
Policy (vs.\ MLP proposal, $N{=}1$) & DR difference [95\% CI] & FQE-direct difference [95\% CI] \\
\midrule
MLP Best-of-8, filtered   & $+1.361$ [$-0.017$, $+1.749$] & $+0.128$ [$+0.123$, $+0.133$] \\
MLP Best-of-8, unfiltered & $+1.369$ [$-0.034$, $+1.754$] & $+0.185$ [$+0.179$, $+0.193$] \\
Critic greedy             & $+1.412$ [$+0.025$, $+1.800$] & $+0.101$ [$+0.093$, $+0.110$] \\
Critic softmax            & $+1.618$ [$+0.063$, $+2.041$] & $+0.034$ [$+0.026$, $+0.043$] \\
\bottomrule
\end{tabular}
\end{adjustbox}
\end{table}

\textbf{Proposal model class.} The MLP proposal had an FQE-direct estimate
of $0.583$, the same point estimate as the supervised language proposal
(Table~\ref{tab:ablation}). Both supervised proposals had negative DR
estimates with wide intervals at $N=1$. For the MLP proposal, the trajectory
ESS was $2.46$ even though the policy was fitted to clinician actions, which
indicates that trajectory-weight degeneracy arises even for imitation
policies: small per-step departures from the estimated behaviour model
compound over the horizon.

\textbf{Value-guided selection.} Best-of-8 selection increased the
FQE-direct estimate of the MLP proposal by $+0.128$ ($[+0.123,+0.133]$) with
the disagreement preference and by $+0.185$ ($[+0.179,+0.193]$) without it.
The corresponding DR intervals narrowly included zero. The selection
operator therefore changed estimated value in the same direction for the
numerical and language proposals under FQE-direct. For the numerical
proposal, the disagreement preference gave a lower FQE-direct estimate than
unfiltered selection; we therefore regard the preference as a conservative
design option rather than as a source of estimated value.

\textbf{Direct use of the critic.} Critic greedy reached an FQE-direct
estimate ($0.684$) similar to that of proposal-constrained selection, but
$91.79\%$ of its stepwise log ratios were clipped and its trajectory ESS was
$1.09$. Its target therefore departs from logged clinician actions at most
decision points, and its value estimates rely more heavily on FQE
extrapolation. Critic softmax had a lower FQE-direct estimate ($0.617$).
Proposal-constrained Best-of-8 selection with the MLP proposal reached
FQE-direct estimates of $0.711$--$0.769$ with $47.71$--$48.34\%$ clipping.
Sampling from a behaviour-supervised proposal before value-based ranking
required fewer clipped stepwise importance ratios than maximisation over
all actions, without guaranteeing adequate state-specific data coverage.

Taken together, these controls show that the selection interface does not
rely on a language proposal outperforming a numerical one. The role of the
language proposal in \method\ is to make an LLM treatment recommender an
explicit policy that this interface can improve and evaluate.

\subsection{Independent-site cohort}

The source-trained MLP proposal was frozen and received source-domain
preprocessing of the shared features. Its 25-action joint distribution was
first marginalised over fluid to five vasopressor levels, as for the
language proposal, before fold-local selection and OPE. Local selection
critics, behaviour models, action thresholds and evaluation models were
fitted on the outer-training
folds under the three-fold cross-fitting protocol of the main text. The
overlap constraint was calibrated within each outer-training partition under
the same protocol and applied to all three control policies before FQE and
DR. Bootstrap resampling was stratified jointly by outer fold and terminal
outcome. ESS is reported as the minimum, over folds and time steps, of the
ESS of the constrained importance weights; all evaluated policies satisfied
the support checks of the adaptation protocol.

\begin{table}[t]
\caption{\textbf{Frozen numerical-proposal control in the independent-site
cohort ($n=571$).} Pooled out-of-fold estimates under the site-adaptation
protocol, with fluid marginalisation before five-level vasopressor selection
and evaluation using overlap-constrained probabilities. Intervals are
95\% conditional patient-bootstrap intervals
stratified by outer fold and terminal outcome.}
\label{tab:ctrl_site}
\centering
\setlength{\tabcolsep}{4pt}
\begin{adjustbox}{max width=\linewidth}
\begin{tabular}{@{}l c c c@{}}
\toprule
Policy & DR [95\% CI] & FQE-direct [95\% CI] & Min.\ fold/time ESS \\
\midrule
Frozen MLP proposal ($N{=}1$)
  & $+0.330$ [$+0.265$, $+0.395$] & $+0.240$ [$+0.222$, $+0.259$] & $97.40$ \\
Frozen MLP Best-of-8, filtered
  & $+0.340$ [$+0.281$, $+0.406$] & $+0.216$ [$+0.198$, $+0.234$] & $173.43$ \\
Frozen MLP Best-of-8, unfiltered
  & $+0.327$ [$+0.265$, $+0.389$] & $+0.223$ [$+0.204$, $+0.240$] & $174.20$ \\
\midrule
\multicolumn{4}{@{}l}{\textit{Paired difference vs.\ frozen MLP proposal ($N{=}1$)}} \\
\addlinespace[2pt]
Frozen MLP Best-of-8, filtered
  & $+0.010$ [$-0.046$, $+0.062$] & $-0.024$ [$-0.034$, $-0.015$] & --- \\
Frozen MLP Best-of-8, unfiltered
  & $-0.003$ [$-0.057$, $+0.051$] & $-0.017$ [$-0.026$, $-0.008$] & --- \\
\bottomrule
\end{tabular}
\end{adjustbox}
\end{table}

At the independent site, Best-of-8 selection did not increase the
FQE-direct estimate of the frozen numerical proposal, and the DR differences
had intervals containing zero (Table~\ref{tab:ctrl_site}). These estimates
describe the frozen numerical proposal under local adaptation.

\section{Implementation settings}
\label{app:hparams}

The numerical architecture settings below describe the internal experiments
with 25 joint actions. Independent-site numerical components use the
five-level vasopressor action space of the adaptation protocol.

The primary language backbone is Qwen3-4B and the within-family scale analysis uses
Qwen3-14B. Models are operated in BF16. The Qwen3-4B proposal is fitted with completion-only LoRA supervision using rank 16, AdamW, learning rate $10^{-4}$, batch size 8, maximum sequence length 512, and 800 training iterations. The proposal uses a single current
observation ($H_{\mathrm{prop}}=1$).

The Qwen3-14B scale-analysis proposal uses the same structured action interface,
LoRA rank, optimiser, learning rate, batch size, maximum sequence length, and
$H_{\mathrm{prop}}=1$ setting, and is trained for 400 iterations. The Qwen3-4B rationale model is fitted separately from the proposal model. It uses LoRA rank 16, AdamW, learning rate $10^{-4}$, batch size 2, maximum
sequence length 1600, and 1300 training iterations. Its input contains at most
four past-only observations.
The proposal and rationale parameters are therefore separate. Only the
proposal distribution enters action selection and off-policy evaluation.

The three-member selection critic uses a two-layer, width-256 LayerNorm
encoder with twin 25-action $Q$ heads and a scalar state-value head. Training
uses AdamW (learning rate $3\times10^{-4}$, weight decay $10^{-4}$, batch size
512), at most 150 epochs with validation patience 25, Polyak rate $0.005$,
Huber $\beta=1.0$, and gradient-norm cap 10. The remaining selection settings
are given in Section~\ref{sec:impl}. In the internal experiments, the
validation-pair disagreement quantile of 0.75 yielded
$\xi\approx0.03623329$ for the frozen ensemble.

The clinician-behaviour model is a 25-action LayerNorm MLP with two hidden
layers of width 256. It is trained using AdamW with learning rate
$3\times10^{-4}$, weight decay $10^{-4}$, batch size 512, at most 80 epochs,
and validation patience 12.
Neural SARSA-Q uses the same numerical encoder and action space and is trained
with logged next actions. Its Boltzmann temperature is selected on validation
data by matching mean action entropy to the clinician-behaviour model rather
than by optimising test-set policy value.

For each target policy, three independently initialised FQE networks are used.
Each is a two-layer LayerNorm MLP with hidden width 256 and a 25-action output.
FQE training uses learning rate $10^{-3}$, weight decay $10^{-4}$, batch size
512, at most 800 epochs, validation patience 60, Polyak rate $0.01$, Huber
parameter $\beta=1.0$, and gradient-norm cap 10. Predictions from the three
models are averaged before computing either FQE-direct or doubly robust
estimates.

\section*{Acknowledgements}
The authors acknowledge Peking University Third Hospital for support in funding and data access. This work was supported by the Peking University Third Hospital Fund for Interdisciplinary Research (No. BYSYJC2024006).

\section*{Declaration of generative AI in scientific writing}
ChatGPT (OpenAI, San Francisco, CA, USA) and Claude (Anthropic) were used to assist with manuscript organisation and English language refinement
under the authors' supervision. The authors review the content and retain
responsibility for the final manuscript.

\section*{Ethics statement}
This retrospective study was approved by the Peking University Third Hospital Medical Science Research Ethics Committee (approval No. (2023) Yilunshen No. 563-02; project No. M2023608). The requirement for informed consent was waived by the Ethics Committee due to the retrospective nature of the study and the use of de-identified clinical data in the hospital cohort. In addition, access to the MIMIC-III database was granted through PhysioNet after completion of the required human subjects research training and credentialing process and acceptance of the corresponding data use agreement. All methods were performed in accordance with relevant guidelines and regulations.

\bibliographystyle{cas-model2-names}
\bibliography{cas-refs}

\end{document}